\documentclass[conference]{IEEEtran}
\usepackage{times}

\usepackage[numbers]{natbib}
\usepackage{multicol}
\usepackage{amsmath}
\usepackage{amssymb}
\usepackage{subcaption}
\usepackage{placeins}
\usepackage{dsfont}
\usepackage[%
  vlined,
  boxed
]{algorithm2e}                          
\usepackage{upgreek}						

\usepackage{graphicx}

\usepackage{color}
\usepackage{xspace}
\usepackage[normalem]{ulem}

\usepackage{todonotes}

\newcommand{\Hypothesis}[2]{ {\setlength{\parindent}{\paperwidth} \textbf{ \noindent H{#1}}\quad\emph{#2} } }

\begin{document}

\title{From the Lab to the Desert: Fast Prototyping and Learning of Robot Locomotion}




%
\author{\authorblockN{Kevin Sebastian Luck\authorrefmark{1}\authorrefmark{4},
Joseph Campbell\authorrefmark{1}\authorrefmark{4},
Michael Andrew Jansen\authorrefmark{2}\authorrefmark{4}, 
Daniel M. Aukes\authorrefmark{3} and
Heni Ben Amor\authorrefmark{1}}
\authorblockA{\authorrefmark{1}School of Computing, Informatics, and Decision Systems Engineering,\\}
\authorblockA{\authorrefmark{2}School of Life Sciences,\\}
\authorblockA{\authorrefmark{3}The Polytechnic School,\\ Arizona State University,
Tempe, Arizona 85281\\ Email: $\lbrace$ksluck, jacampb1, majanse1, danaukes, hbenamor$\rbrace$@asu.edu}
\authorblockA{\authorrefmark{4}Authors contributed equally}}

\maketitle

\begin{abstract}
We present a methodology for fast prototyping of morphologies and controllers for robot locomotion. Going beyond simulation-based approaches, we argue that the form and function of a robot, as well as their interplay with real-world environmental conditions are critical. Hence, fast design and learning cycles are necessary to adapt robot shape and behavior to their environment. To this end, we present a combination of laminate robot manufacturing and sample-efficient reinforcement learning. We leverage this methodology to conduct an extensive robot learning experiment. Inspired by locomotion in sea turtles, we design a low-cost crawling robot with variable, interchangeable fins. Learning is performed using both bio-inspired and original fin designs in an artificial indoor environment as well as a natural environment in the Arizona desert. The findings of this study show that static policies developed in the laboratory do not translate to effective locomotion strategies in natural environments. In contrast to that, sample-efficient reinforcement learning can help to rapidly accommodate changes in the environment or the robot. 
\end{abstract}

\IEEEpeerreviewmaketitle

\section{Introduction}
Robots are often tasked with operating in challenging environments that are difficult to model accurately. Search-and-rescue or space exploration tasks, for example, require robots to navigate through loose, granular media of varying density and unknown composition, such as sandy desert environments. A common approach is to use simulations in order to develop ideal locomotion strategies before deployment. Such an approach, however, requires prior knowledge about ground composition which may not be available or may fluctuate significantly. In addition, the sheer complexity of such terrain necessitates the use of approximations when simulating interactions between the robot and its environment. However, inaccuracies inherent to approximations can lead to substantial discrepancies between simulated and real-world performance. These limitations are especially troublesome as robot design is also guided by simulations in order to overcome time constraints and material deterioration associated with traditional physical testing. 
 
In this work we argue that the design of effective locomotion strategies is dependent on the interplay between (a) the shape of the robot, (b) the behavioral and adaptive capabilities of the robot, and (c) the characteristics of the environment. In particular, adverse and dynamic terrains require a design process in which both \emph{form} and \emph{function} of a robot can be rapidly adapted to numerous environmental constraints. To this end, we introduce a novel methodology employing a combination of fast prototyping and manufacturing with sample-efficient reinforcement, thereby enabling practical, physical testing-based design.

\begin{figure}[t!]
	\centering
    \includegraphics[width=0.45\textwidth]{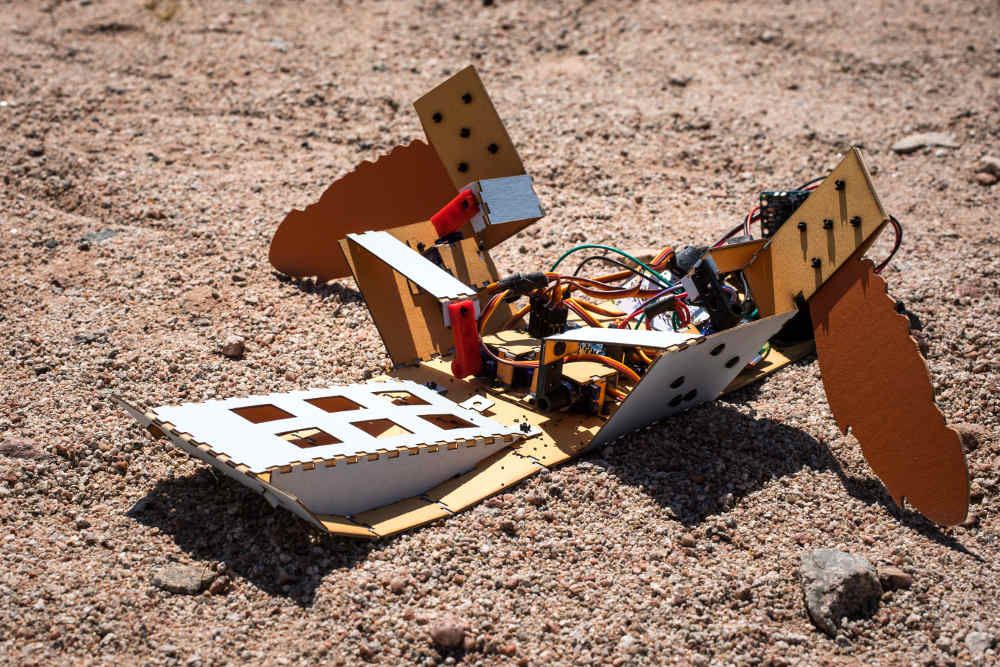}
	\caption{A robot made from a multi-layer composite learns how to move across sand in the Arizona desert.} 
	\label{img:teaser}
\end{figure}

First, we describe a manufacturing process in which \emph{foldable robotic devices} (Fig. \ref{img:teaser}) are constructed out of a \emph{single planar shape} consisting of multiple laminated layers of material. The overall production time of a robot using this manufacturing method is in the range of a few hours, i.e., from the first laser-cut to the deployment. As a result, changes to the robot shape can be performed by quickly iterating over several low-cost design cycles.

In addition to rapid design refinement and iteration, the synthesis of effective robot control policies is also of vital importance. Variations in terrain, the assembly process, motor properties, and other factors can heavily influence the optimal locomotion policy. Manual coding and adaptation of control policies is, therefore, a laborious and time-intensive process which may have to be repeated whenever the robot or terrain properties change, especially drift in actuation or changes in media granularity. Reinforcement learning (RL) methods~\cite{Sutton:1998:IRL:551283} are a potential solution to this problem. Using a trial-and-error process, RL methods explore the policy space in search of solutions that maximize the expected reward, e.g., the distance traveled while executing the policy. However, RL algorithms typically require thousands or hundreds of thousands of trials before they converge on a suitable policy~\cite{mnih-dqn-2015}. Performing large numbers of experiments on a physical robot causes wear-and-tear on hardware, leads to drift in sensing and actuation, and may require extensive human involvement. This severely limits the number of learning experiments that can be performed within a reasonable amount of time.

A key element of our approach is a sample-efficient RL~\cite{luck2016sparse} method which is used for swift learning and adaptation whenever the changes occur to the robot or the environment. By leveraging the low-dimensional nature and periodicity of locomotion gaits, we can rapidly synthesize effective control policies that are best adapted to the current terrain. We show that using this method, the learning process quickly converges towards appropriate policy parameters. This translates to learning times of about 2-3 hours on the physical robot.  

We leverage this methodology to conduct an extensive robot learning experiment. Inspired by locomotion in sea turtles, we design a low-cost crawler robot with variable, interchangeable fins. Learning is performed with different bio-inspired and original fin designs in both an indoor, artificial environment, as well as a natural environment in the Arizona desert. The findings of this experiment indicate that artificial environments consisting of poppy seeds, plastic granulates or other popular loose media substitutes may be a poor replacement for true environmental conditions. Hence, even policies that are not learned in simulation, but rather on granulate substitutes in the lab may not translate to reasonable locomotion skills in the real-world. In addition, our findings show that reinforcement learning is a crucial component in adapting and coping with variability in the environment, the robot, and the manufacturing process.  

We thus demonstrate that the combination of a rapid prototyping process for robot design (form) and the fast learning of robot policies (function) enables environment-adaptive robot locomotion. 

\section{Related Work}
\label{sec:related_work}
Prior studies have indicated that locomotion in granulate media is dependent upon successful compaction of the substrate, without causing fluidization \cite{li2013,mazouchova2013,mazouchova2010}. Unfortunately, the dynamic response of granulate media during locomotion is difficult to predictively simulate or replicate \cite{li2013,mazouchova2013,askari2016}. In desert environments, this difficulty is compounded by the \textit{heterogeneous} composition of the loose, sandy topsoil, making it nearly impossible to predict the effectiveness of any locomotor strategy \textit{a priori} \cite{li2013,askari2016}. In practice, the performance of robotic systems in heterogeneous granulate media, particularly in xeric habitats, must be evaluated \textit{post-hoc} and iteratively improved (see for example \cite{mazouchova2013}) through successive design refinements and adaptive learning of locomotion. 

Finned animals, and sea turtles in particular, have achieved highly stable and efficient locomotion through heterogeneous granulate substrata \cite{mazouchova2010,li2013,mazouchova2013}. Of the many animals capable of effective locomotion in sand, we drew most heavily upon the sea turtle due to the simplicity and stability of its motion \cite{wyneken1997}. Unlike sand-swimming animals, like sand lizards~\cite{maladen2009undulatory}, finned animals (such as sea turtles) require fewer degrees of freedom and actuated joints to achieve forward motion. 

A robotic analogue to sea turtles, FlipperBot (FBot), was designed to provide a two-limbed approximation of sea turtle locomotion in an ongoing effort to characterize the motion of finned animals through sand \cite{mazouchova2013}. Unlike other robotic devices inspired by turtles, FBot was designed for locomotion in granulate media and not for swimming \cite{low2007modular, yao2013development}. FBot features two degrees of freedom for each limb; however, the fins were configured such that they could be either fixed (relative to the arm) or free to rotate. The combined quasi-static motion of the limbs was similar to a ``breast stroke'', dragging the body through the sand \cite{mazouchova2013}. 

In general, the ``bio-inspired robotics'' approach~\cite{bhushan2009biomimetics} has proven fruitful for designing laboratory robots with new capabilities (new gaits, morphologies, control schemes), including rapid running \cite{clark2001biomimetic,plaAbue06}, slithering \cite{tesch2009parameterized}, flying \cite{ma2013controlled}, and ``swimming" in sand \cite{maladen2009undulatory}. In addition, using the biologically inspired robots as ``physical models'' of the organisms has revealed scientific insights into the principles that govern movement in biological systems, as well as new insights into low-dimensional dynamical systems (see for example~\cite{holAful06} and references therein).
Our work differs fundamentally from these works, not only in execution, but also in principle: we aim to generate optimal motion through bio-mimicry and learning, rather than learning how optima are generated in a biological system.

\section{Methodology}

In this section, we describe our methodology for fast robot prototyping and learning. We discuss a sample-efficient reinforcement learning method that enables fast learning of new locomotion skills. In combination with a laminate robot manufacturing process, our method allows for rapid iterations over both form and function of a robot. The main rationale behind this approach is that environmental conditions are often hard to reproduce outside of the natural application domain. Hence, the development cycle should be informed by experiences in the real application domain, e.g., on challenging terrain such as desert environments. Our approach facilitates this process and significantly reduces the underlying development time. Consequently, we will describe the methodology for prototyping form and function in more detail.  

\begin{figure*}[ht!]
    	\centering
    	\begin{subfigure}[b]{0.29\textwidth}
        	\centering
            \includegraphics[height=0.75\textwidth]{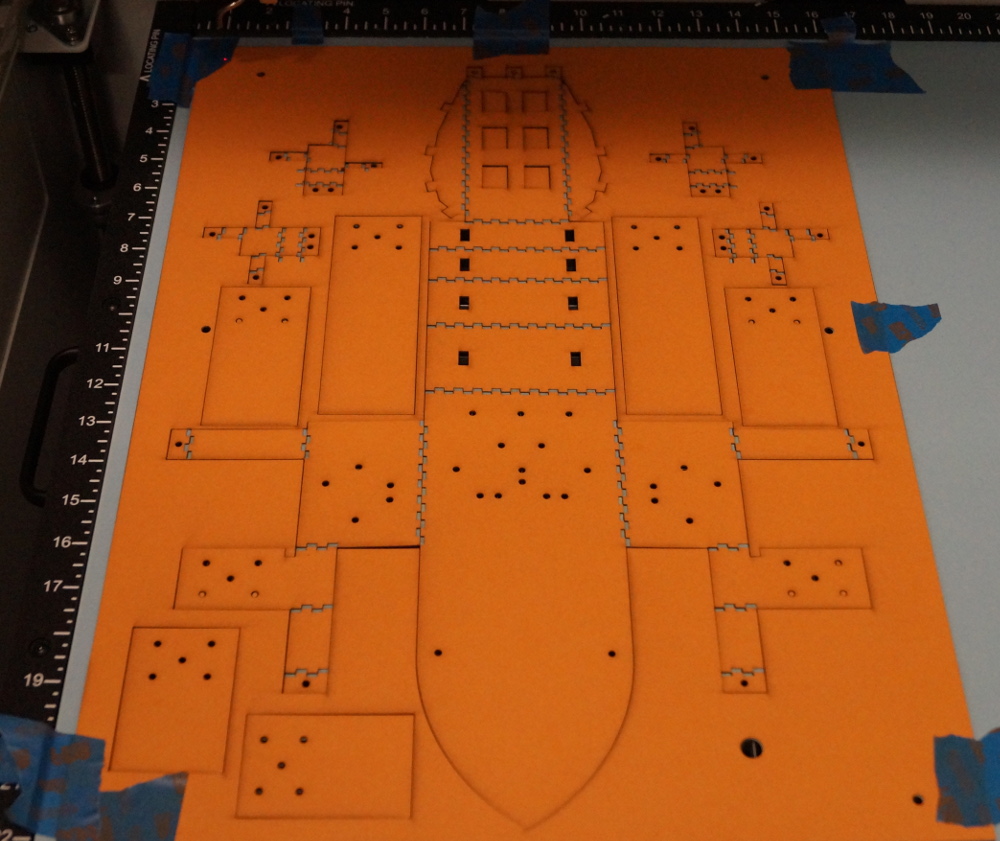}
            \caption{}
            \end{subfigure}~%
        \begin{subfigure}[b]{0.29\textwidth}
        	\centering
            \includegraphics[height=0.7\textwidth]{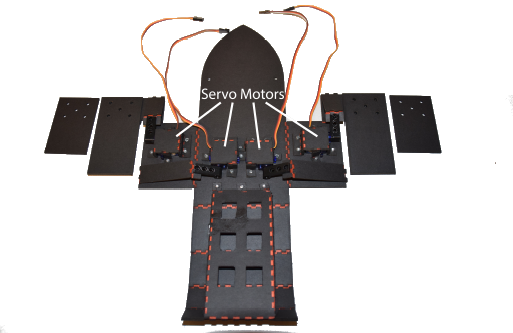}
            \caption{}
            \end{subfigure}~%
        \begin{subfigure}[b]{0.4\textwidth}
        	\centering
            \includegraphics[width=1\textwidth]{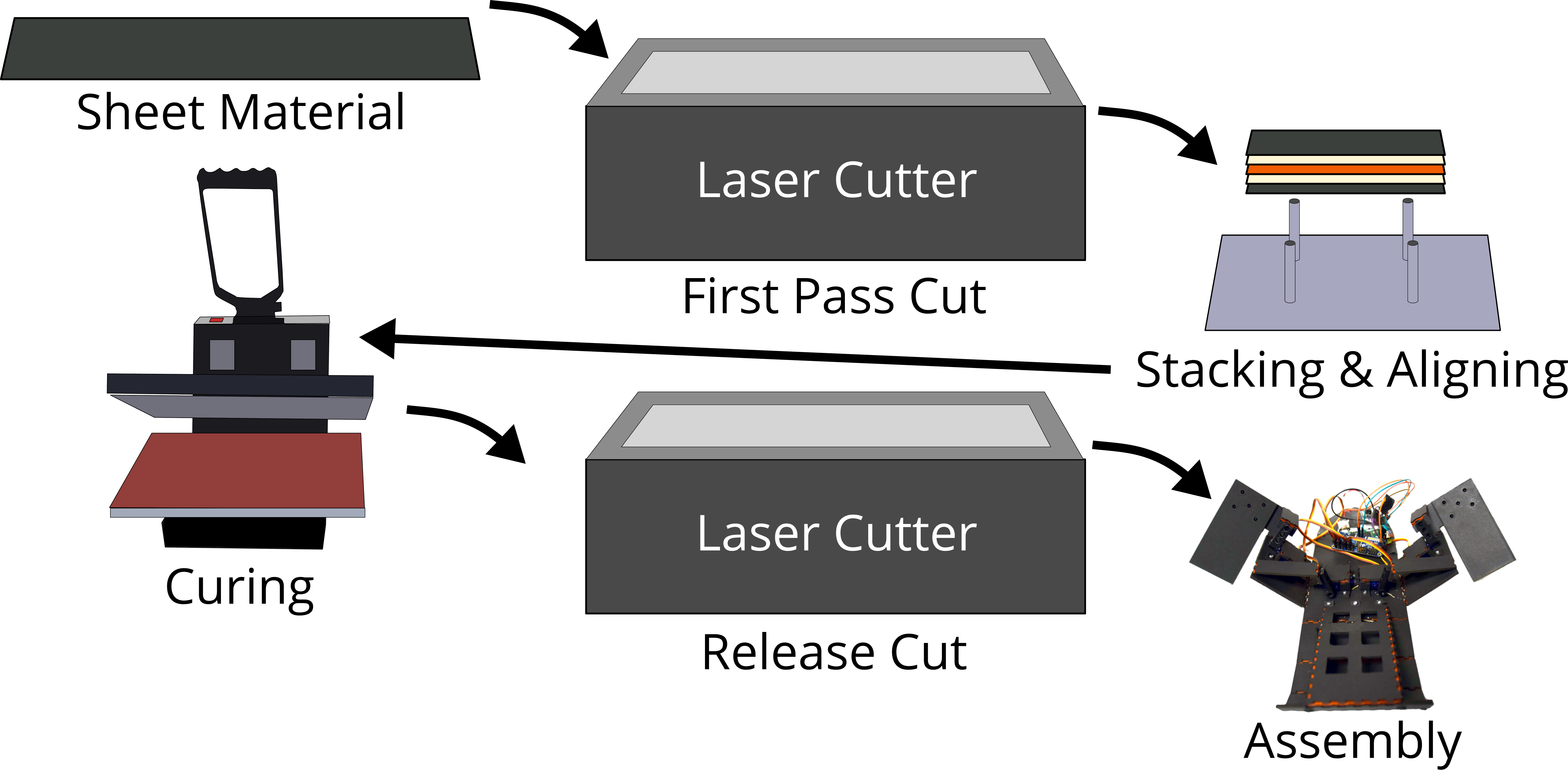}
            \caption{}
            \label{img:fab-process}
            \end{subfigure}~%
            \caption{Manufacturing of laminate robotic mechanisms: (a) The robot components are engraved using a laser cutter on planar sheets and later laminated. (b) The components are folded into a robotic structure. The motors, the control board, and the battery are added manually. (c) The full fabrication process.\label{fig:sheets}}
\end{figure*}

\subsection{FORM: Laminate Robot Manufacturing}
Laminate manufacturing can be used in order to construct affordable, light-weight robots. Laminate fabrication processes (known as SCM~\cite{Wood2008}, PC-MEMS~\cite{Sreetharan2012}, popup-book MEMS~\cite{Whitney2011}, lamina-emergent mechanisms~\cite{Gollnick2011}, etc.) permit rapid construction from planar sheets of material which are iteratively cut, aligned, stacked, and laminated to form a composite material.

\begin{figure}[htb]
	\centering
	\includegraphics[width=.33\textwidth]{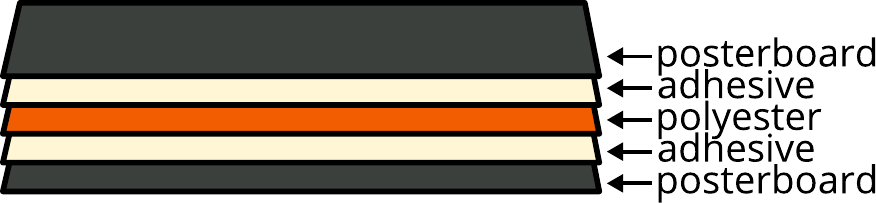}
	\caption{The laminates involved are constructed as a sandwich of five layers: poster-board, adhesive, polyester, adhesive, and poster-board.}
	\label{img:laminate}
\end{figure} 
 
The laminate mechanisms discussed in this paper were printed in five layers.  As shown in Figure \ref{img:laminate}, two rigid layers of 1mm-thick poster-board are sandwiched around two adhesive layers of Drytac MHA heat-activated acrylic adhesive, which is itself sandwiched around a single layer of 50 $\mu$m-thick polyester film from McMaster-Carr. Flat sheets of each material are cut out on a laser cutter, then stacked and aligned using a jig with holes pre-cut in the first pass of the laser cut.  Once aligned, these layers are fused together using a heated press in order to create a single laminate.  The adhesive cures at around 85-104 degrees Celsius, and comes with a paper backing which allows it to be cut, aligned with the poster-board, and laminated.  The paper backing is then removed, and the two adhesive-mounted poster-board layers are aligned with the center layer of polyester and laminated again.  This laminate is returned to the laser, where a second release cut permits the device to be separated from surrounding scrap material and erected into a final three-dimensional configuration. 

Laminate mechanisms resulting from this process are capable of a high degree of precision through bending of flexure-based hinges created through the selective removal of rigid material along desired bend axes.  With fewer rolling contacts(bearings) than would typically be found in traditional robots, laminate mechanisms are ideal in sandy environments, where sand infiltration can shorten service life.  Connections between layers can be established through adhesive layers, in addition to plastic rivets which permit quick attachment between laminates.  Mounting holes permit attaching a variety of off-the-shelf components including motors, micro-controllers, and sensors. Rapid attachment/detachment is a highly desired feature for this platform, as different flipper designs can be tested using the same base platform.  In all, this fabrication method permits \emph{rapid iteration} during the design phase, and \emph{rapid re-configuration} for testing a variety of designs across a wide range of force and length scales, due to its compatibility with a wide range of materials.  Fig.~\ref{fig:sheets}(a) depicts the basic planar sheets after cutting. Fig.~\ref{fig:sheets}(b) shows the individual components of the robot after they are detached from the sheets and folded into a structure. Fig.~\ref{fig:sheets}(c) is the full fabrication process.  The whole manufacturing process of one robot takes up to 50 minutes while the 3D-printing process of four horns, which serve as connections between the motors and the paper, takes 58 minutes.

\subsection{FUNCTION: Sample-Efficient Reinforcement Learning}
In this section we discuss a sample-efficient RL method that converges on optimal locomotion policies within a small number of robot trials. Our  approach leverages two key insights about human and animal locomotion. In particular, locomotion is (a) inherently low-dimensional and based on a small number of motor synergies~\cite{Krouchev1991}, as well as (b) highly periodic in nature.

\begin{figure}
    	\centering
    	\begin{subfigure}[b]{0.22\textwidth}
        	\centering
            \includegraphics[width=\textwidth]{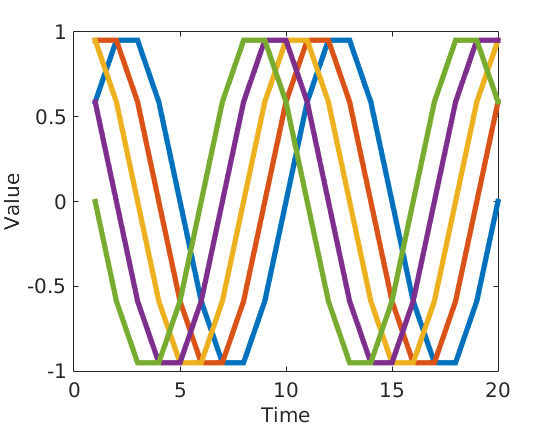}
            \caption{Basis functions $\upphi(t)_{1:5}$.}
        \end{subfigure}%
        \begin{subfigure}[b]{0.22\textwidth}
        	\centering
            \includegraphics[width=\textwidth]{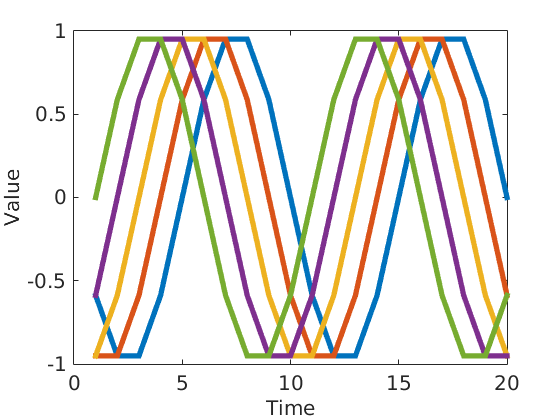}
            \caption{Basis functions $\upphi(t)_{6:10}$.}
        \end{subfigure}
        \caption{The sinusoidal basis functions $\upphi(t)$ used for learning in this paper. Each basis function is based on a sine curve and shifted in time. The final policy is based on a linear combination of these functions.}
        \label{img::basisfunctions}
    \end{figure}

To implement these insights within a reinforcement learning framework, we build upon the Group Factor Policy Search (GrouPS) algorithm introduced by Luck et al.~\cite{luck2016sparse}. GrouPS jointly searches for a low-dimensional control policy as well as a projection matrix $\mathbf{W}$ for embedding the results into a high-dimensional control space. It was previously shown~\cite{luck2016sparse} that the algorithm is able to uncover optimal policies after a few iterations with only hundreds of samples. Group Factor Policy Search models the joint actions as $\mathbf{a}_t^{(m)} = (\mathbf{W}^{(m)}\mathbf{Z} + \mathbf{M}^{(m)} + \mathbf{E}^{(m)})\mathbf{\upphi}(\mathbf{s},t)$ for each time step t of a trajectory and  each $m$-th group of actions. The matrix $\mathbf{W}$ represents the transformation matrix from the low dimensional to the high dimensional space (exploitation) and $\mathbf{M}$ the parameters of the current mean policy. The entries of the matrices $\mathbf{Z}$ and $\mathbf{E}$ are Gaussian distributed with $z_{ij} \sim N(0,1)$ for the latent values and $e_{ij} \sim N(0,\tau_m^{-1})$ for the isotropic exploration. The function $\mathbf{\upphi}(\mathbf{s},t)$ consists of basis functions $\mathbf{\upphi}_i(\mathbf{s},t)$ and depends in our experiments only of the time step $t$ and not of the full state $\mathbf{s}$. In contrast to the work in~\cite{luck2016sparse}, however, we incorporate periodicity constraints into the search process by focusing on periodic feature functions. 
We use periodic basis functions over 20 time steps for the control signal, see Fig.~\ref{img::basisfunctions}. Given a point in time $t$ we compute each control dimension $a_i$ by
\begin{equation}
    a_i = \sum_j \left( \tilde{w}_{ij} + m_{ij} + e_{ij} \right) \sin\left( \frac{t}{T} 720^{\circ} + \frac{j - 1}{J} 360^{\circ}\right)
\end{equation}
with $ \tilde{w}_{ij} = \sum_k w_{ik} z_{kj} $ and $J$ being the number of basis functions in $\mathbf{\upphi}(s,t)$.

GrouPS also takes prior information about potential groupings of joints into account when searching for an optimal transformation matrix $\mathbf{W}$. For our robotic device we used two groups: the first group consists of the two fin-joints and the second group of the two base-joints. Thus, we exploit the symmetry of the design. The number of dimensions of the manifold was set to three and the rank parameter, controlling the sparsity and structure of $\mathbf{W}$, to one. The outline of the algorithm can be found in Algorithm 1. 
Incorporating dimensionality reduction, periodicity, and information about the group structure yields a highly sample-efficient algorithm.

    {

  \newcommand{\ivec}[1]{\ensuremath{\boldsymbol{\mathrm{#1}}}}
  \newcommand{\imat}[1]{\ensuremath{\boldsymbol{\mathrm{#1}}}}
  
  \newcommand{\idistributed}{\ensuremath{\sim}}						
  \newcommand{\iidentitymat}{\imat{I}}								
  \newcommand{\itrajectory}{\ensuremath{\cvec{\uptau}}}				
  \newcommand{\istpolicy}{\ensuremath{\pi}}							
  
  \newcommand{\itrans}{^\mathrm{T}}									
  \newcommand{\iinv}{^{-1}}											
  \newcommand{\idiff}[2]{\frac{\partial #2}{\partial #1}}				
  \newcommand{\inormaldist}[2]{\mathcal{N} \left( #1,#2 \right)}		
  \newcommand{\inormaldistp}[3]{\mathcal{N} \left( #1 \middle\vert #2,#3 \right)}		
  \newcommand{\imatnormaldist}[4]{\mathcal{N}_{#1,#2} \left( #3,#4 \right)} 
   \newcommand{\igammadist}[2]{\mathcal{G} \left( #1,#2 \right)}		
   \newcommand{\igammadistp}[3]{\mathcal{G} \left( #1 \middle\vert #1, #2 \right)}		

  \newcommand{\ikron}{\otimes}										
  \newcommand{\itrace}[1]{\mathrm{tr}\left(  #1 \right)}				
  \newcommand{\iexpect}[1]{\mathds{E}\left[ #1 \right]}	
  \newcommand{\iexpectl}[2]{\mathds{E}_{#2}\left[ #1 \right]}	
  \newcommand{\idiag}[1]{\mathrm{diag}\left( #1 \right)}						

  \newcommand{\parameter}{\ensuremath{\theta}}
  \newcommand{\parameterV}{\ensuremath{\mathbf{\theta}}}
  \newcommand{\iKLD}[2]{\mathsf{KL}\left( #1 \parallel #2 \right)}
  \newcommand{\idx}{\mathrm{d}}
  \newcommand{\ireal}{\mathbb{R}}
  
  \newcommand{\itraj}{\ivec{\tau}}
  \newcommand{\istate}{\ivec{s}}
  \newcommand{\iaction}{\ivec{a}}
  \newcommand{\iW}{\imat{W}}
  \newcommand{\ibasisf}{\ivec{\phi}}
  \newcommand{\imean}{\imat{M}}
  \newcommand{\itau}{\tilde{\tau}}
  \newcommand{\ilatent}{\tilde{\ivec{z}}}
  \newcommand{\inormreward}{\hat{R}(\itraj)}

  \newcommand{\cvec}[1]{\boldsymbol{\mathrm{#1}}}
  \newcommand{\cmat}[1]{\boldsymbol{\mathrm{#1}}}
  
\newcommand{\mmc}[1]{\mathcal{#1}}
\newcommand{\set}[1]{{{\mathcal{#1}}}} 
\newcommand{\transpose}{^\mathrm{T}}
\newcommand{\inverse}{^{-1}}
\newcommand{\diff}[2]{\frac{\partial #2}{\partial #1}}
\newcommand{\traj}{\cvec{\uptau}}
\newcommand{\parameterO}{\uptheta}
\newcommand{\stateS}{s}
\newcommand{\actionS}{a}
\newcommand{\state}{\cvec{s}}
\newcommand{\action}{\cvec{a}}
\newcommand{\policy}{\pi}
\newcommand{\intd}{\text{d}}
\newcommand{\Erw}{\mathds{E}}
\newcommand{\Cov}{\text{Cov}}
\newcommand{\paraM}{{\cmat{M}}}
\newcommand{\param}{{\cvec{m}}}
\newcommand{\paraW}{{\cmat{W}}}
\newcommand{\paraw}{{\cvec{w}}}
\newcommand{\paralatentTwo}{{\ensuremath{\tilde{\cvec{z}}}}}
\newcommand{\paralatentTWO}{{\ensuremath{\tilde{\cmat{Z}}}}}
\newcommand{\paratau}{{\tilde{\tau}}}
\newcommand{\normdist}{\mathcal{N}}
\newcommand{\gammadist}{\mathcal{G}}
\newcommand{\feature}{{\cvec{\upphi}}}
\newcommand{\trace}[1]{{\text{Tr}\left( #1 \right)}}
\newcommand{\alphadiag}{\bar{\bar{\cmat{\alpha}}}}
\newcommand{\hyperTauA}{a^{\paratau}}
\newcommand{\hyperTauB}{b^{\paratau}}
\newcommand{\hyperAlphaA}{a^{\alpha}}
\newcommand{\hyperAlphaB}{b^{\alpha}}
\newcommand{\normReward}{\widehat{R}}
\newcommand{\realnumbers}{\mathds{R}}

\SetAlCapSkip{1ex}
\IncMargin{1em}
\begin{algorithm}
  \DontPrintSemicolon
  \KwIn{Reward function $R\left(\cdot\right)$ and initializations of parameters. Choose number of latent dimension $n$ and rank $r$. Set hyper-parameter and define groupings of actions.}
  \hrulefill\\
  \While{reward not converged}{
    \For(\# Sample H rollouts){h=1:H}{
      \For{t=1:T}{
	$\action_t = \paraW \cmat{Z} \feature + \paraM \feature + \cmat{E}\feature$ \\~~with $\cmat{Z} \sim \normdist \left( \cvec{0}, \cmat{I} \right)$ and
	$\cmat{E} \sim \normdist \left( \cvec{0}, \cmat{\paratau} \right)$, where $\cmat{\paratau}^{(m)} = \paratau_m\iinv \cmat{I}$\;
	Execute action $\action_t$\;
      }
      Observe and store reward $R\left(\traj\right)$\;	 	
    }
    \vspace{8px}
    Initialization of q-distribution\;
    \While{not converged}{
      Update $q\left(\paraM\right)$, $q\left(\paraW\right)$, $q\left(\paralatentTWO\right)$, $q\left(\cmat{\alpha}\right)$ and  $q\left(\cvec{\paratau}\right)$\;
    }
    $\paraM = \Erw_{q\left(\paraM\right)}\left[\paraM\right]$\;
    $\paraW = \Erw_{q\left(\paraW\right)}\left[\paraW\right]$\;
    $\cmat{\alpha} = \Erw_{q\left(\cmat{\alpha}\right)}\left[\cmat{\alpha}\right]$\;
    $\cvec{\paratau} = \Erw_{q\left(\cvec{\paratau}\right)}\left[\cvec{\paratau}\right]$\;
  }
  \hrulefill\\
  \KwResult{Linear weights $\paraM$ for the feature vector $\feature$, representing the final policy. The columns of $\paraW$ represent the factors of the latent space.}
  \caption{Outline of the Group Factor Policy Search (GrouPS) algorithm as presented in \cite{luck2016sparse}.
  }
  \label{alg::newAlg}
\end{algorithm}
}

\section{A Foldable Robotic Sea Turtle}
\label{sec:robot_description}

With the general methodology established, this section introduces the design of the robotic device used in this research.
As discussed in Sec. \ref{sec:related_work}, our design takes inspiration from sea turtles. By necessity, the design also conforms to the constraints of the laminate fabrication techniques being employed -- primarily that it is composed of a single planar layer. The salient aspects of Chelonioid morphology integrated into our design are described below.




\subsection{Biological Inspiration}   

The design of our laminate device was primarily inspired by the anatomy and locomotion of sea turtles. We chose to focus on the terrestrial locomotion of \textit{adult} sea turtles, rather than juveniles or hatchlings, emphasizing the greater load-bearing capacity and stability of their anatomy and behavior. There are seven recognized species in Cheloniodea in six genera \cite{pritchard1999}. In spite of considerable inter-specific differences in morphology, all sea turtles use the same set of anatomical features to generate motion. Specifically, adult sea turtles support themselves on the radial edge of the forelimbs to (1) elevate the body (thus reducing or eliminating drag) and (2) generate forward motion \cite{wyneken1997}. This unique behavior allows these large and exceedingly heavy animals (up to 915\,kg in \textit{Dermochelys coriacea} (Vandelli, 1761)\footnote{Pursuant to the International Code of Zoological Nomenclature, the first mention of any specific epithet will include the full genus and species names as a binomen (two part name) followed by the author and date of publication of the name. This is \textit{not} an in-line reference; it is a part of the name itself and refers to a particular species-concept as indicated in the description of the species by that author.}) to move in a stable and effective manner through granular media \cite{eckert1988}. 
        
\begin{figure}
    	\centering
    	\begin{subfigure}[b]{0.22\textwidth}
        	\centering
            \includegraphics[width=\textwidth]{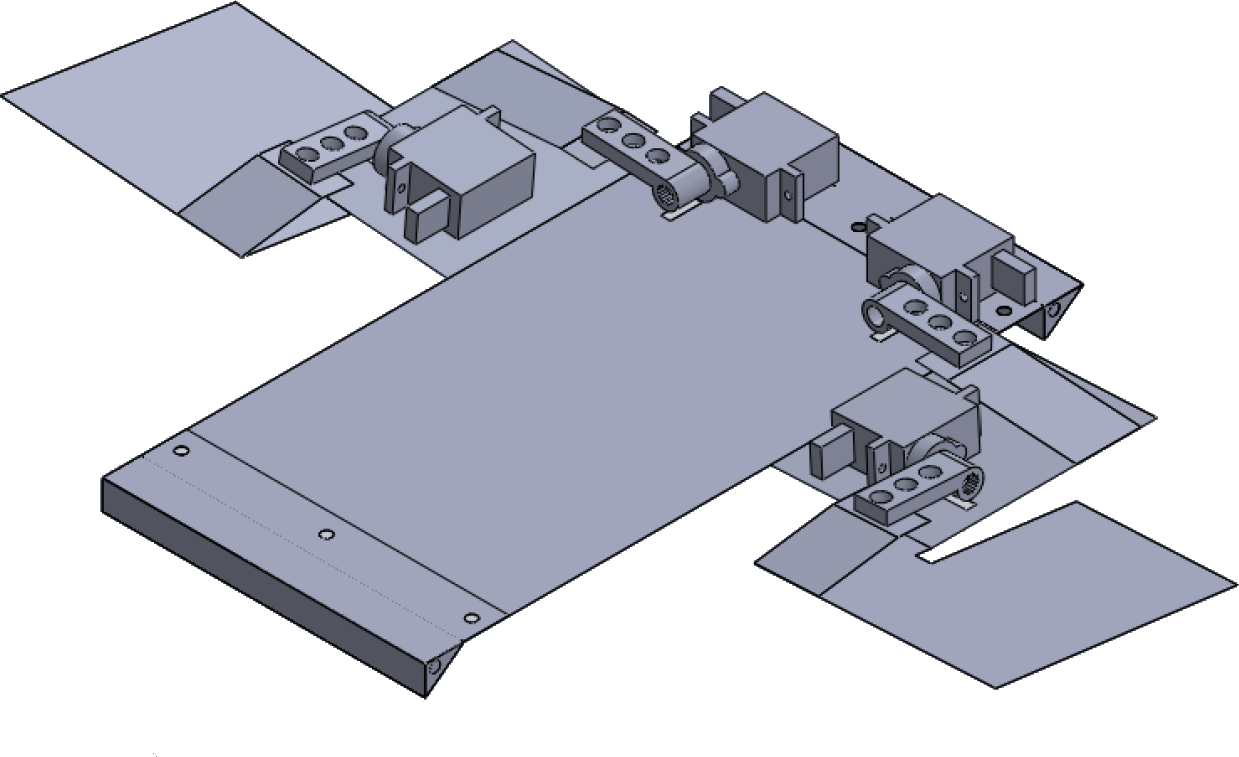}
            \caption{}
        \end{subfigure}%
        \begin{subfigure}[b]{0.22\textwidth}
        	\centering
            \includegraphics[width=\textwidth]{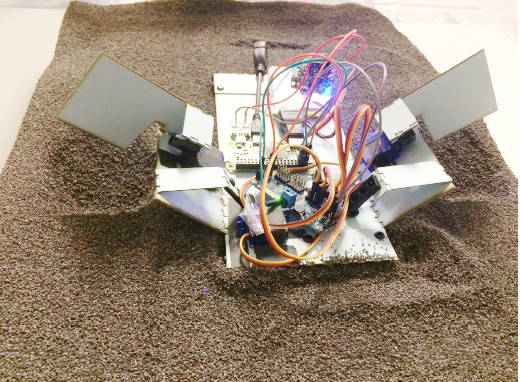}
            \caption{}
        \end{subfigure}
        \caption{The initial ``flat body" design of the robot. The front of the robot buried into the sand during motion. The body was later curved}
        \label{img::initialdesign}
\end{figure}
    
        \begin{figure*}
    	\centering
        \begin{subfigure}[b]{0.225\textwidth}
        	\centering\includegraphics[width=\textwidth]{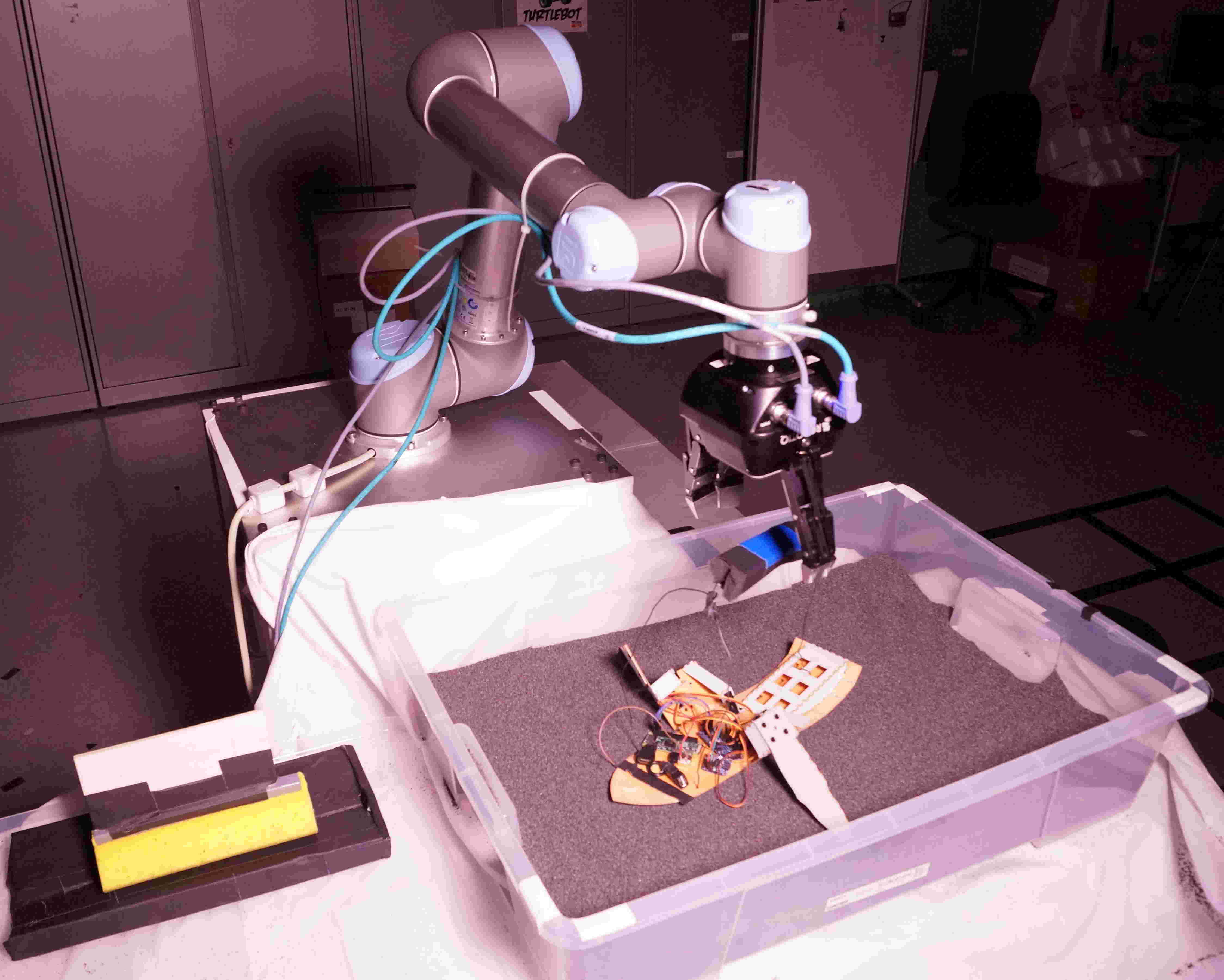}
            \caption{}
        \end{subfigure}~%
        \begin{subfigure}[b]{0.24\textwidth}
        	\centering\includegraphics[width=\textwidth]{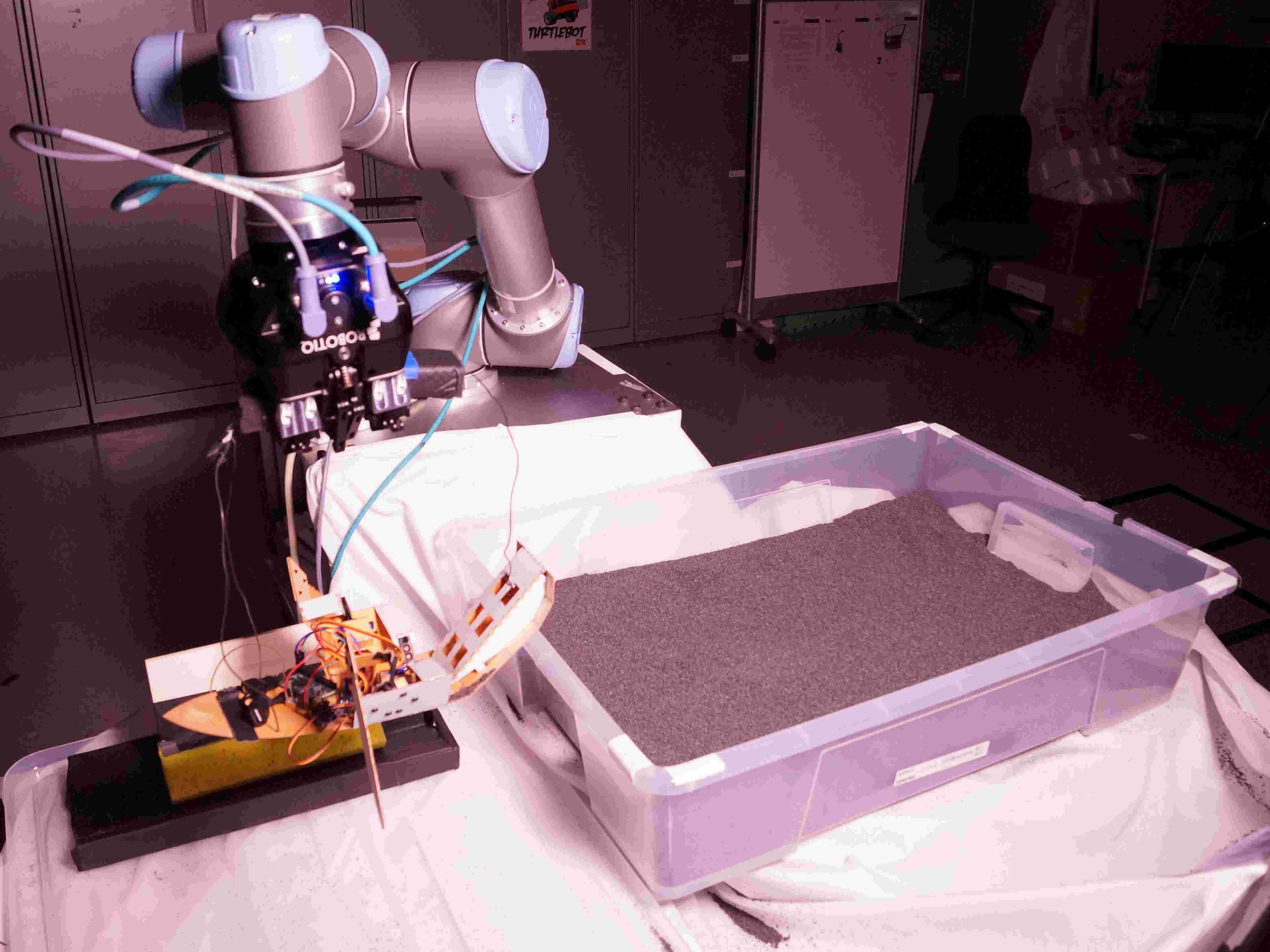}
            \caption{}
        \end{subfigure}~%
        \begin{subfigure}[b]{0.22\textwidth}
        	\centering\includegraphics[width=\textwidth]{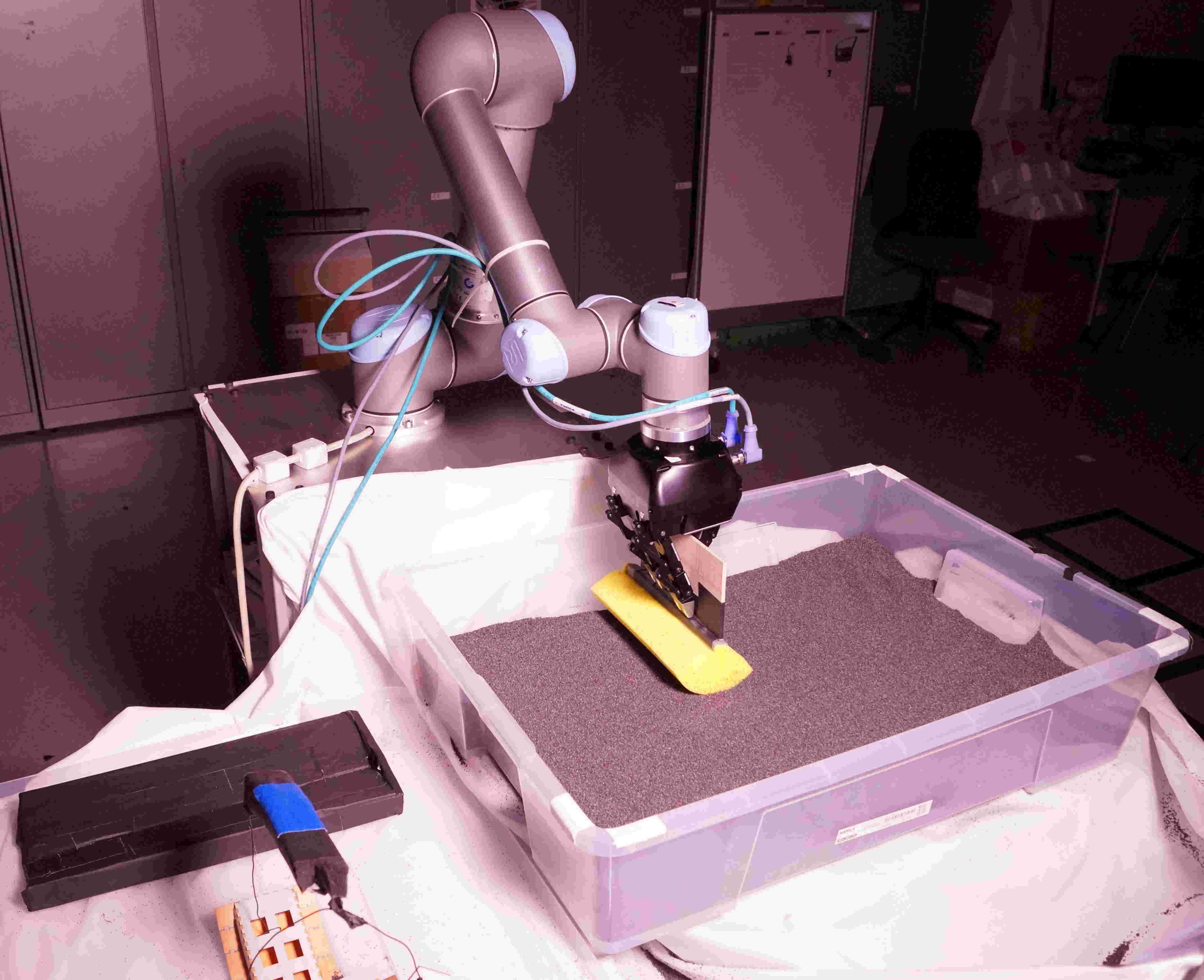}
            \caption{}
        \end{subfigure}~%
        \begin{subfigure}[b]{0.24\textwidth}
        	\centering\includegraphics[width=\textwidth]{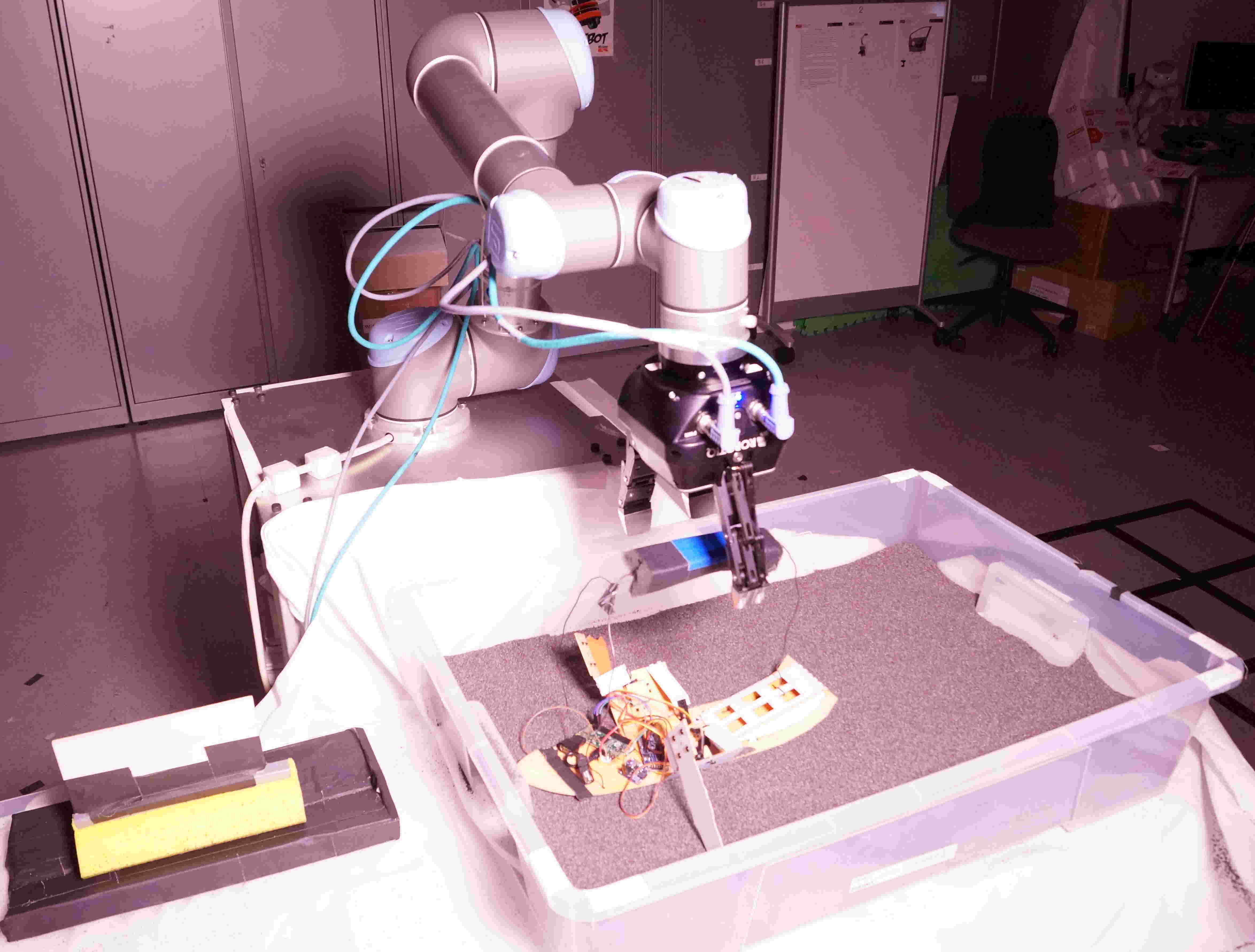}
            \caption{}
        \end{subfigure}%
        \caption{Sequence of actions the robotic arm executes in each learning cycle: (a) First the robot under test is located in the testbed, grasped and then (b) subsequently moved into a resting position. The robotic arm proceeds to (c) smooth the testbed with a tool. Finally, the robot under test is (d) put into its initial position and the next trajectory is executed.}
        \label{img::learning_automation}
    \end{figure*}
    
\subsection{Robot Design}

Focusing on the turtle's forelimb for generating locomotion, the robot form and structure was determined within an iterative design cycle. In all designs, the body was suitably broad to prevent sinking during forward motion, and remained in contact with the ground at rest. This provides stability while removing the need for the limbs to bear the weight of the body at all times. A major benefit of this configuration is that only the two forelimbs are needed to generate forward thrust. Transmission of load occurs primarily under tension (as in muscles), to accommodate the laminate material and to provide dampening to reduce joint wear. The limbs have 2 rotational degrees of freedom, such that the fins move down and back into the substrate, while the body moves up and forward. This two degree of freedom arm was sufficient to replicate the circular motion of the fins (and particularly of the radial edge) observed in sea turtles (see \cite{wyneken1997}). 

Initial experiments attempted on early prototypes revealed a critical design flaw: the anterior end was prone to ``plowing'' into the substrate (see Fig.~\ref{img::initialdesign}). This limitation was solved by mimicking two features of turtle anatomy. First, the apical portion of the design is shaped to elevate the body above sand, with an upturned apex, similar to upturned intergular and gular scales of the anterior sea turtle plastron (see \cite{pritchard1999}). Second, the back end of the body was tapered to reduce drag (as compared to a rectangular end of equal length). 

\begin{figure}
    	\centering
        \includegraphics[width=0.3\textwidth]{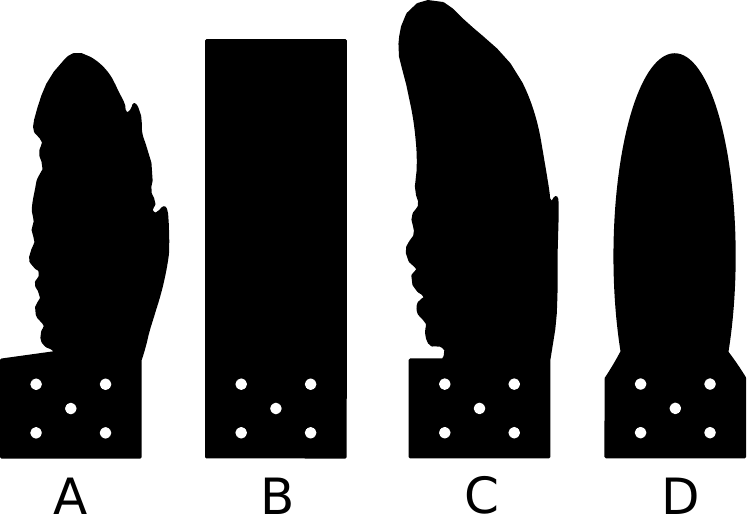}
        \caption{The four different design of fins used for the presented robotic device. Designs A and C are accurate reproductions of the actual shape of sea turtle fins, namely \textit{Caretta caretta} (A) and \textit{Natator depressus} (C). Designs B and D are simple rectangular and ellipsoid shapes.}
\label{img::design::finns}
\end{figure}

In the final design cycle, we also sought to mimic and explore the morphology of the fins. Extant sea turtle species exhibit a variety of fin shapes and include irregularities seen on the outer edges, such as scales and claws. These features are known to be used for terrestrial locomotion by articulating with the surface directly (rather than being buried in the substrate) \cite{mazouchova2010,dodd1988}. In order to understand how fin shape affects locomotion performance, we designed four pairs of fins: two generated from outlines of sea turtle fins which include all irregularities (\textit{Caretta caretta} (Linnaeus, 1758) and \textit{Natator depressus} (Garman, 1880), from \cite{pritchard1999}), and two based on artificial shapes with no irregularities, as shown in Fig.~\ref{img::design::finns}. All of these were attached to the main body at a position equivalent to the anatomical location of the humeroradial joint (part of the elbow in the fin), and scaled to the width of the body. The arms of the robot were designed such that the fins can be interchanged at will, allowing for easy comparison of fin performance.

\section{Experiments}
In this section, we focus on evaluating the locomotion performance of the prototypes generated with our laminate fabrication process. In particular, the robustness to variations stemming from the terrain and manufacturing process, and the sensitivity to changes in the physical fin shape.
%
%

    	\begin{figure*}[htb]
    	\centering
    	\begin{subfigure}[b]{0.32\textwidth}
        	\centering
            \includegraphics[width=\textwidth]{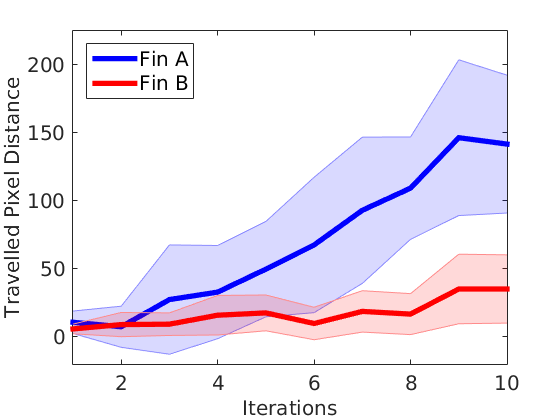}
            \caption{Comparison between fin A and fin B.}
            \label{img::exp::exhaustive::ab}
        \end{subfigure}~%
        \begin{subfigure}[b]{0.32\textwidth}
        	\centering            
            \includegraphics[width=\textwidth]{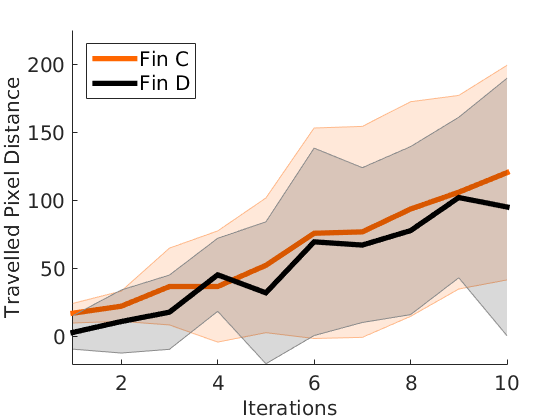}
            \caption{Comparison between fin C and fin D.}
        \end{subfigure}
        \begin{subfigure}[b]{0.32\textwidth}
        	\centering
            \includegraphics[width=\textwidth]{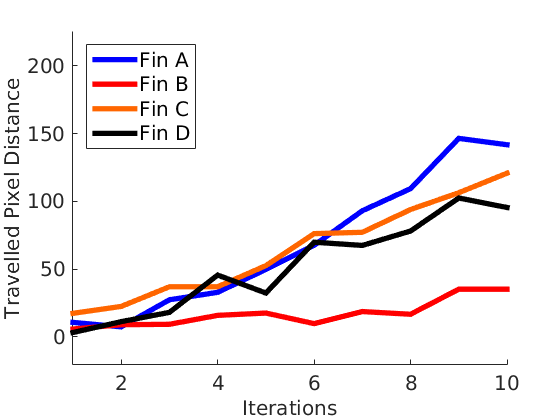}
            \caption{Comparison between all fins.}
        \end{subfigure}
        \caption{Comparison between the learning for different fin designs. Each experiment was performed five times and mean/standard deviations were computed. The learning process was performed on poppy seeds.}
        \label{img::exp::exhaustive}
    \end{figure*}

More formally, there are three hypotheses that we experimentally evaluate:

\Hypothesis{1}{Group Factor Policy Search is able to find an improved locomotion policy -- with respect to distance traveled forward -- in a limited number of trials, despite the presence of variations in the rapidly prototyped robotic device and the environment.}
\\
\Hypothesis{2}{The shape of the fin influences the performance of the locomotion policy.}
\\
\Hypothesis{3}{The locomotion policies learned in the natural environment out-perform those learned in the artificial environment, when executed in the natural environment.}

These hypotheses are tested through the following experiments.



\subsection{Evaluation of Fin Designs}

    
    

This experiment is designed to evaluate the effectiveness of locomotion policies generated for the four types of fins described in Sec.~\ref{sec:robot_description}.
Five independent learning sessions were conducted for each fin, consisting of 10 policy search iterations each for a total of 1050 policy executions per fin.
The experiment was performed in an indoor, artificial environment utilizing poppy seeds (similar to \cite{mazouchova2013flipper}) as a granulate material substitute for sand -- they are less abrasive and increase the longevity of prototypes. Human involvement, and thus randomness, was minimized during the learning process by employing an articulated robotic arm (UR5). This arm was responsible for placing the robot under test in the artificial environment prior to each policy execution, then subsequently removing it and resetting the environment with a leveling tool. This sequence of actions is depicted in Fig.~\ref{img::learning_automation}.

The policy search reward was automatically computed by measuring the distance (in pixel values) that a target affixed to the robot traveled with a standard 2D high-definition webcam.
This was computed from still frames captured before and after policy execution.
After learning, the mean iteration policies were manually executed and measured in order to produce metric distance rewards for comparison.

    \subsection{Policy Learning in a Desert Environment}
    
\begin{figure}[tb]
    	\centering
        \includegraphics[width=0.25\textwidth]{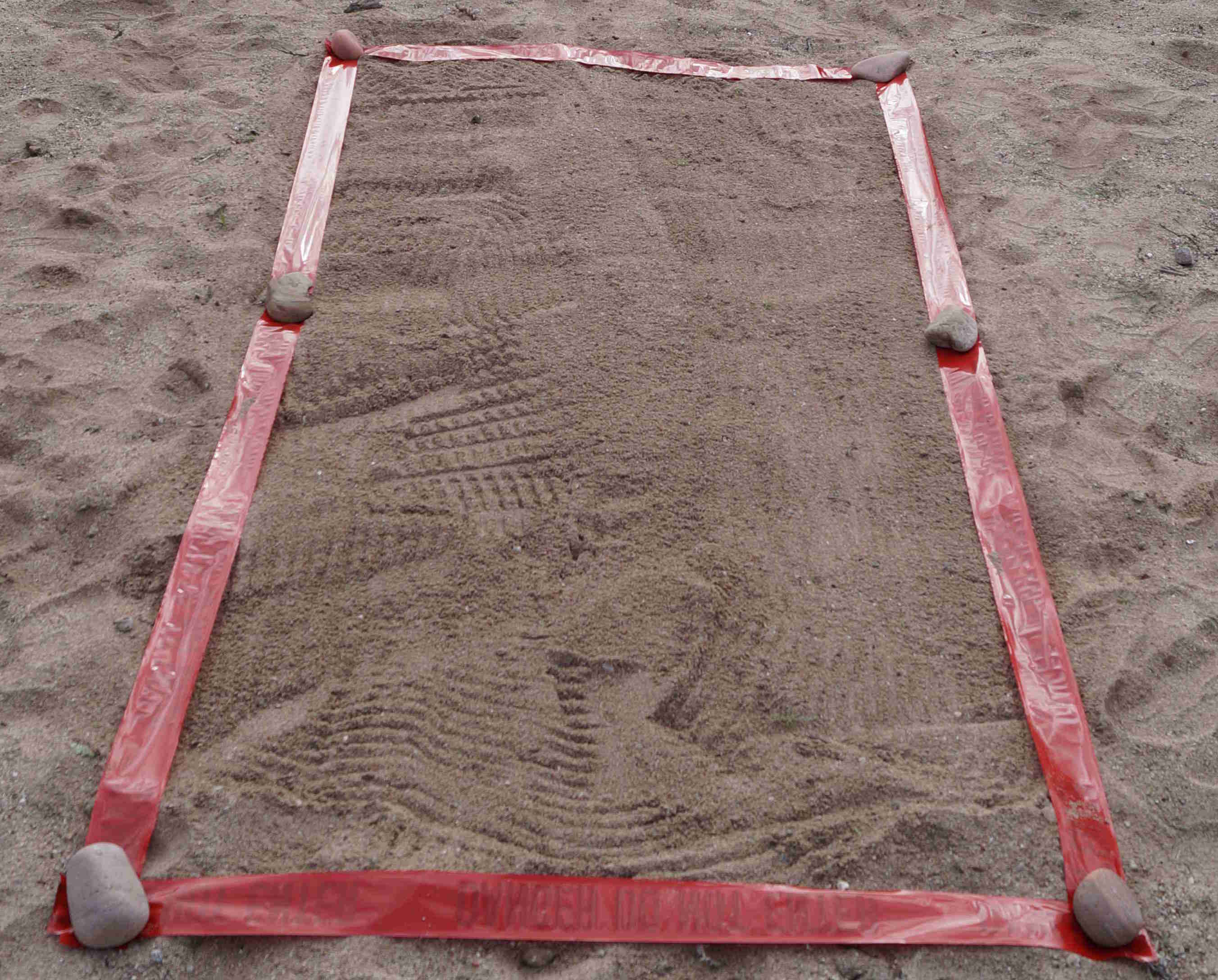}
        \caption{The testbed in the Arizona desert used for evaluating and learning policies in a real environment. The surface of the testbed is flattened in order to increase comparability between the values measured for each policy.}
        \label{img::testbed}
    \end{figure}

The second experiment was designed to test how well policies transfer between environments, and whether policies learned in-situ are more effective than policies learned in other environments.
Over the course of two days, the policies generated for each fin in the artificial environment from the first experiment were executed 
in a desert environment in the Tonto National Forest of Arizona
in order to measure their distance rewards.
We opted to create a flattened test bed as shown in Fig.~\ref{img::testbed}, rather than using untouched ground, in order to reduce locomotion bias due to inclines, rocks, and plants.

Furthermore, two additional learning sessions were conducted for fins A and C in the same test bed in order to provide a point of comparison.
To maintain consistency with the first experiment, learning was performed with 10 Policy Search iterations and reward values were measured via camera.
Manually measured distance values for each mean iteration policy were obtained after learning.
A video of the learning process and supplementary material can be found on \url{http://www.c-turtle.org}.

\section{Results}

\begin{figure}[htb]
    	\centering
        \includegraphics[width=0.35\textwidth]{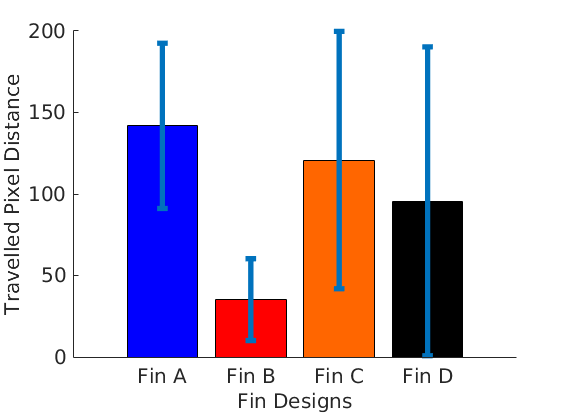}
        \caption{The mean and standard deviation of policies for each fin design in the last iteration of the learning process. The rewards represent the distance the robot moved forward.}
        \label{img::lastiteration}
    \end{figure}

	The rewards achieved by policies learned on poppy seeds are presented in Figure \ref{img::exp::exhaustive} with their mean and standard deviation over the conducted experiments. Figure \ref{img::exp::exhaustive} (a) compares the biologically inspired fin A (\textit{C. caretta}) and the simple rectangular shape. The second biologically inspired fin C (\textit{N. depressus}) and the artificial oval fin can be found in Figure \ref{img::exp::exhaustive} (b), both with a similar performance. The mean values of the learned policies are given in Figure \ref{img::exp::exhaustive} (c). The reward in these plots is given as pixel distances, as recorded by the camera, covered by the robot with its movement, which means that fin A (C. caretta) outperforms all other fin designs. On the opposite, the rectangular shaped fin shows the worst performance.
	This can also be seen in Figure \ref{img::lastiteration} which compares the mean and standard deviation of achieved rewards in the last iteration of the learning process between the four different fin designs.

        \begin{figure}[htb]
        \centering
    	\begin{subfigure}[b]{0.5\textwidth}
        	\centering
            \includegraphics[width=0.7\textwidth]{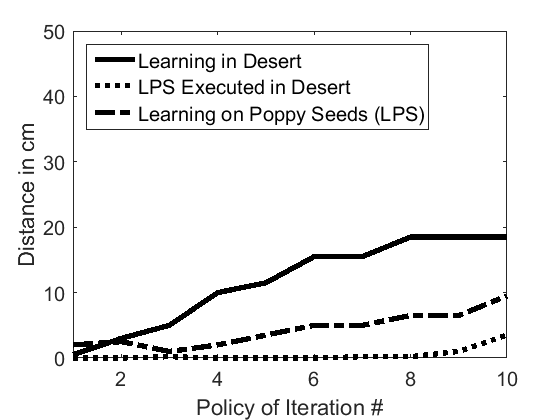}
            \caption{Comparison between policies learned for fin design C. The algorithm was initialized with the same random number generator for learning.}
        \end{subfigure}
        \begin{subfigure}[b]{0.5\textwidth}
        	\centering
            \includegraphics[width=0.7\textwidth]{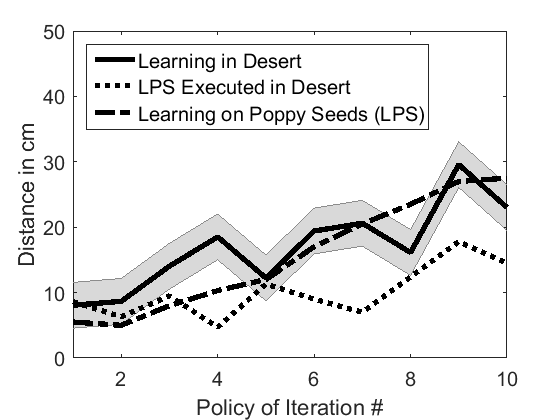}
            \caption{Comparison between policies learned for fin design A. The algorithm was initialized with the same random number generator for learning. Due to a technical issue only pixel distances were recorded for learning in the desert. For comparability those pixel distances were transformed into centimeters but are attached with a variance of about 3.5cm.}
        \end{subfigure}
        \caption{Comparison between polices learned on poppy seeds and executed on poppy seeds (LPS), learned on poppy seeds and executed in a desert environment, and policies learned and executed in a desert environment.}
        \label{img::exp::outdoor}
    \end{figure}
    
    Two different fin designs, A (\textit{C. caretta}) and C (\textit{N. depressus}), were selected for the comparison between policies learned on poppy seeds and policies learned in a natural environment. Figure \ref{img::exp::outdoor} (a) and (b) show the covered distances in centimeters for policies learned  and executed on poppy seeds as well as executed in the desert for each iteration. The third policy for each fin was learned and evaluated in the desert. It can be seen that the policy learned in the natural environment outperforms the policies learned on the substitute in the laboratory environment. 
    
	\begin{figure*}
    
		\centering
   		\begin{subfigure}[b]{\textwidth}
        	\centering
        	\includegraphics[width=0.19\textwidth]{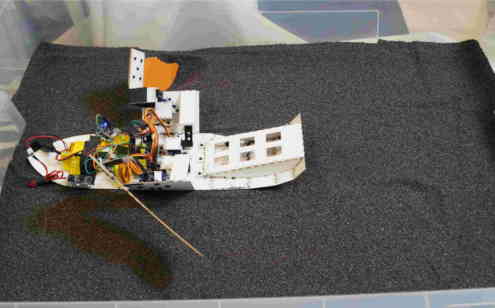}~%
            \includegraphics[width=0.19\textwidth]{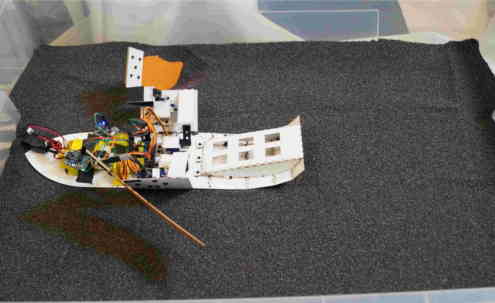}~%
            \includegraphics[width=0.19\textwidth]{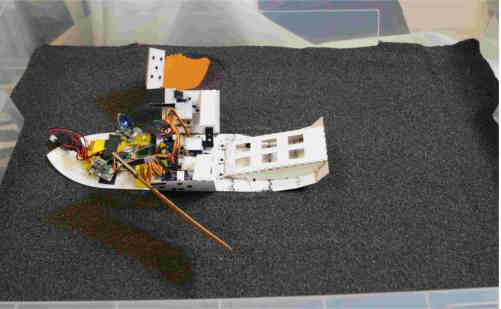}~%
            \includegraphics[width=0.19\textwidth]{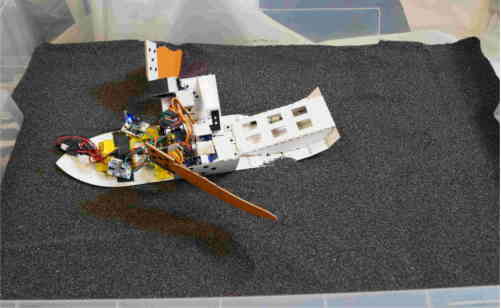}~%
            \includegraphics[width=0.19\textwidth]{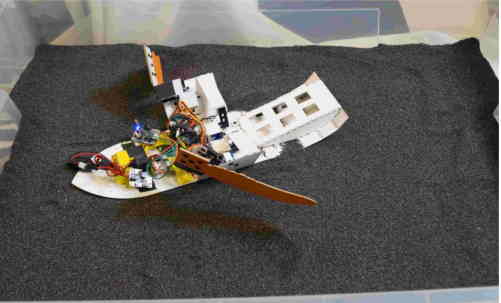}
        	\caption{Executions of policies learned on poppy seeds. The start position of the robot was on the wall of the testbed on the left side. \hspace*{9ex}}
    	\end{subfigure}
        
        \begin{subfigure}[b]{\textwidth}
        	\centering
        	\includegraphics[width=0.19\textwidth]{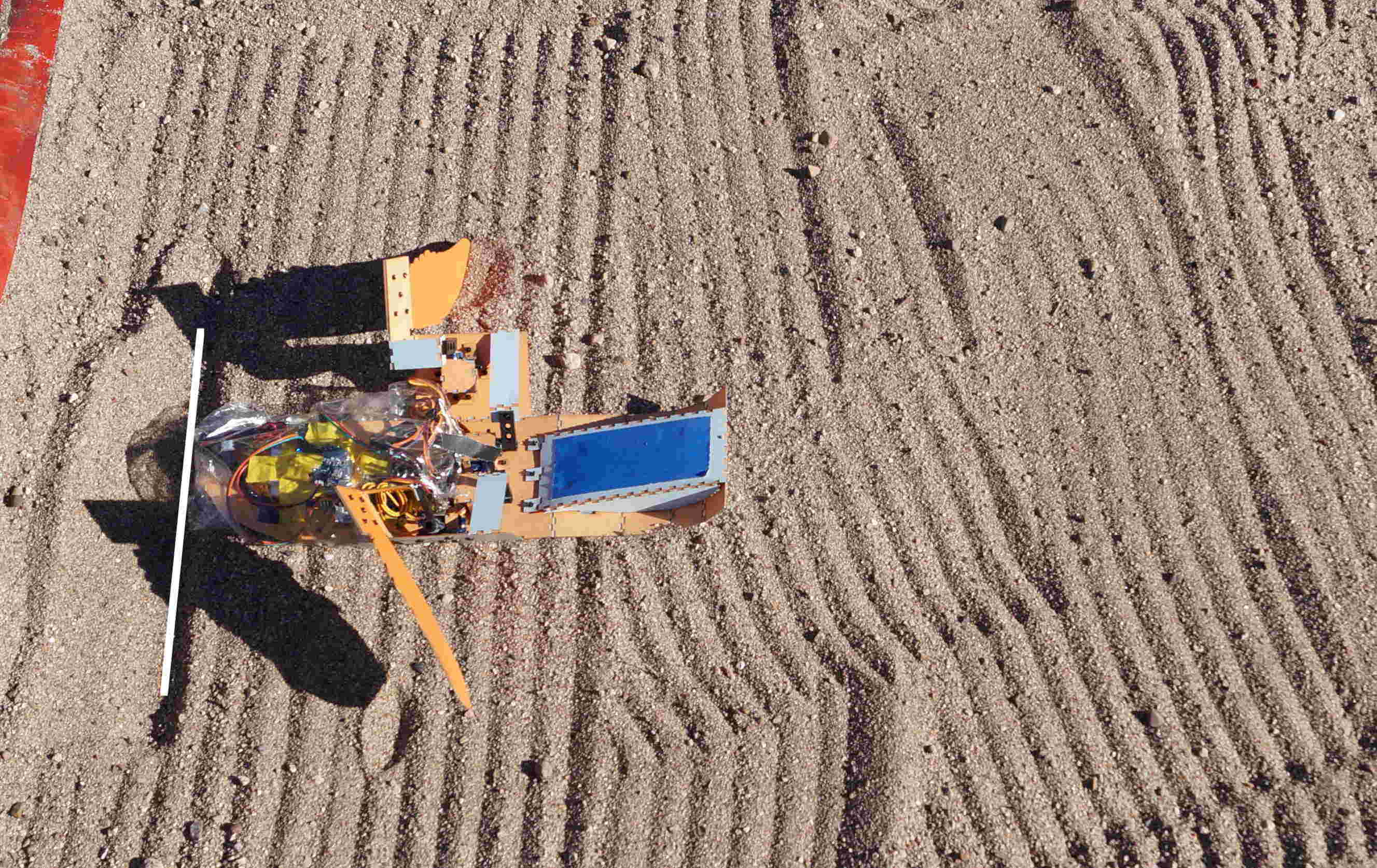}~%
            \includegraphics[width=0.19\textwidth]{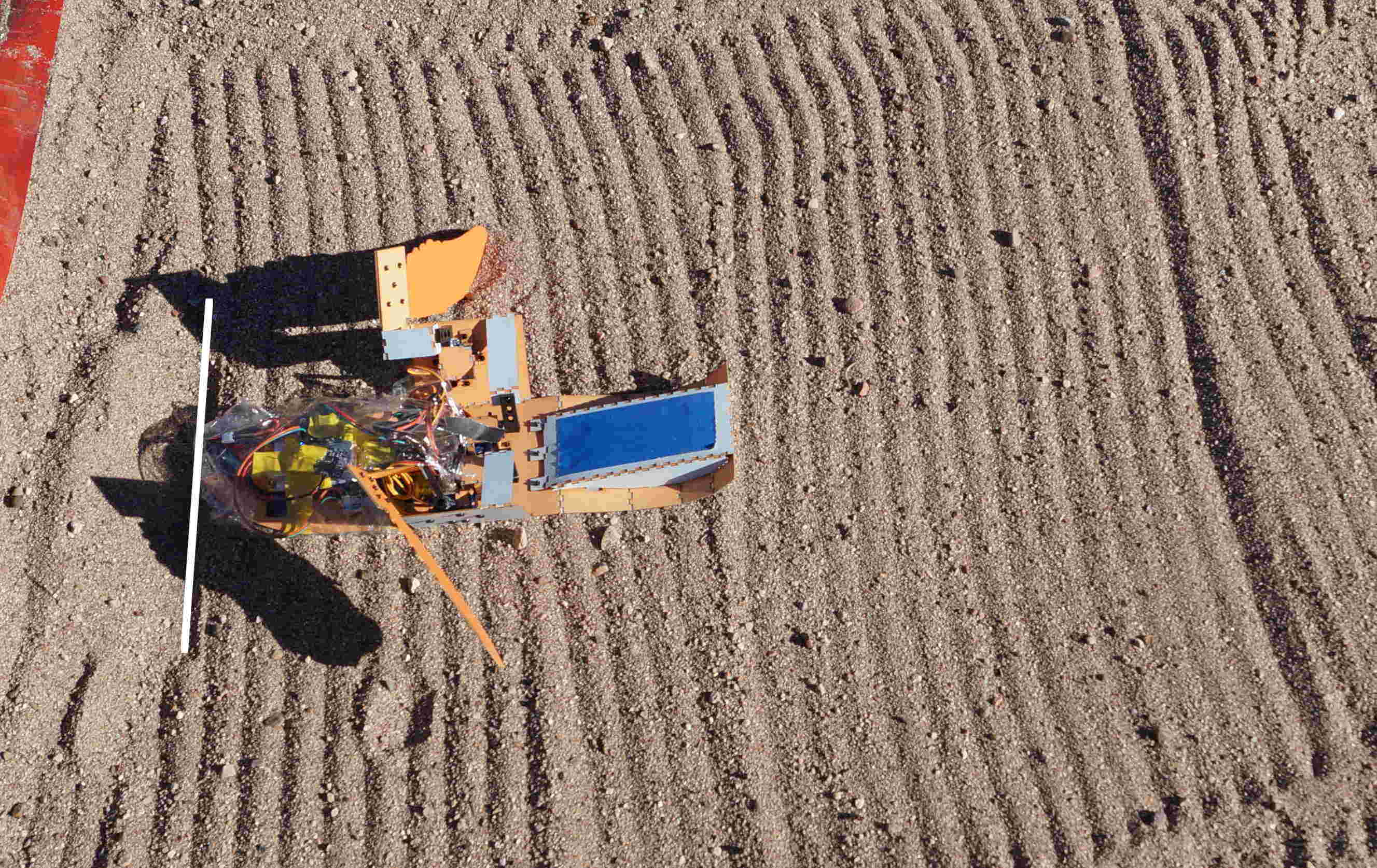}~%
            \includegraphics[width=0.19\textwidth]{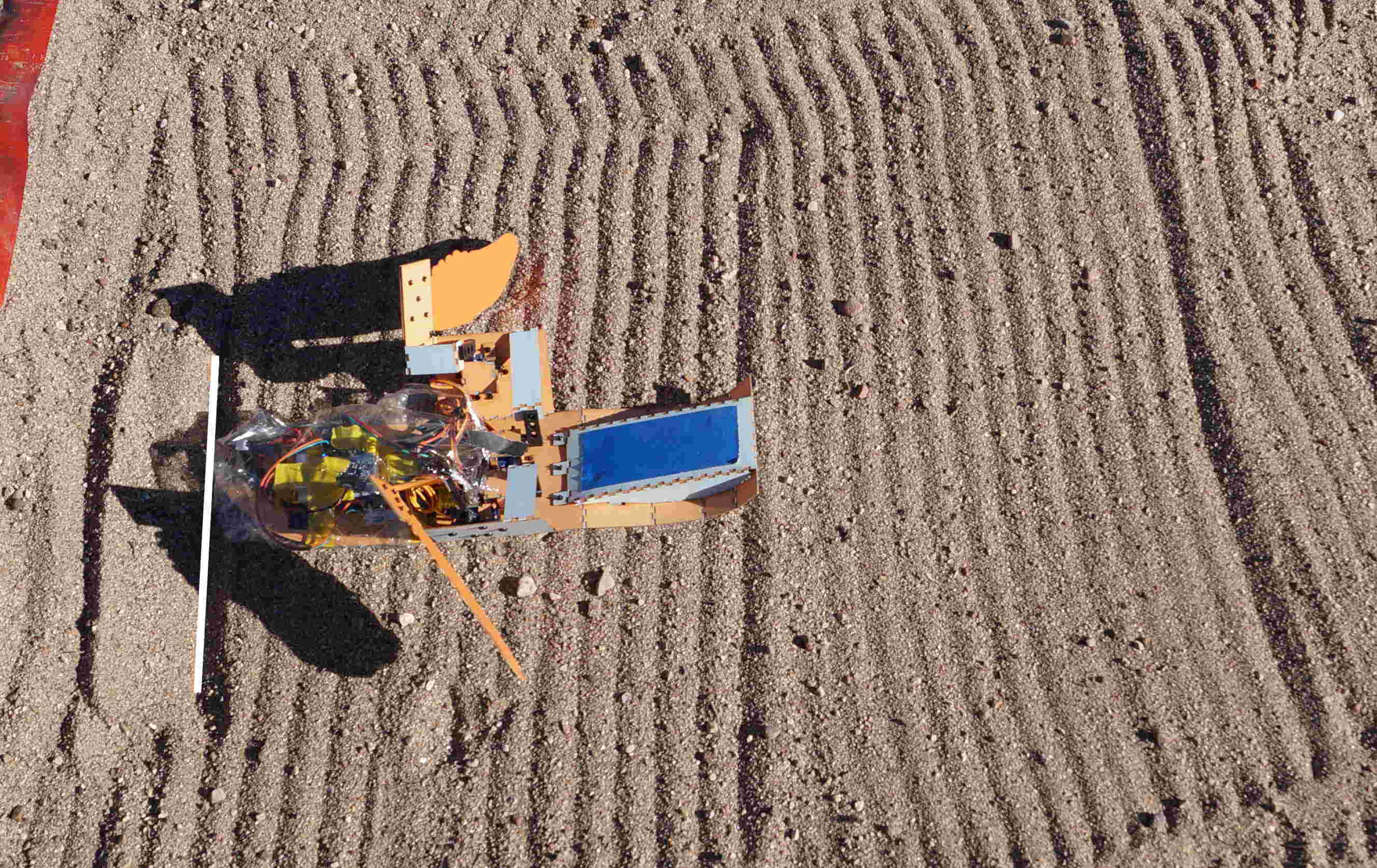}~%
            \includegraphics[width=0.19\textwidth]{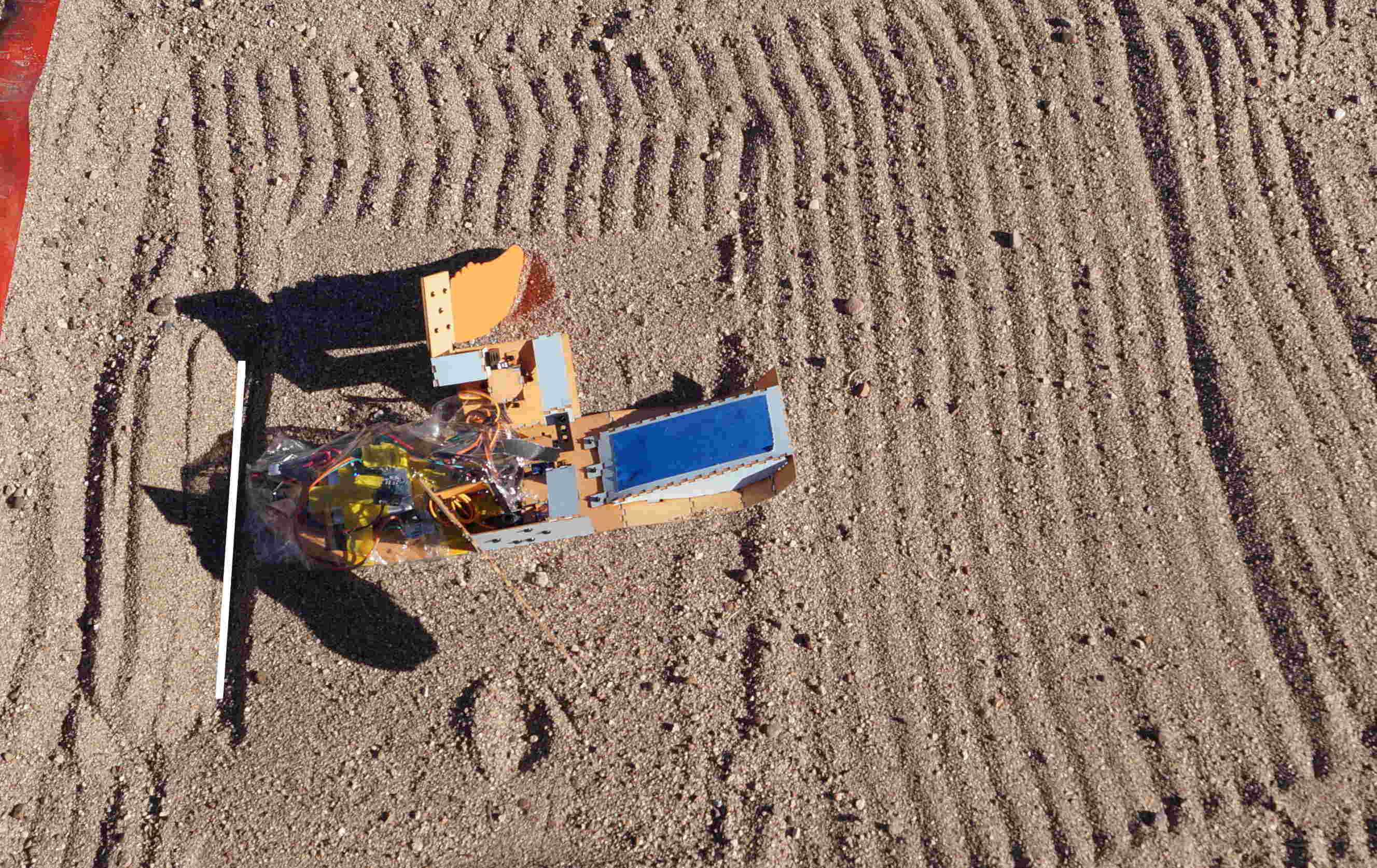}~%
            \includegraphics[width=0.19\textwidth]{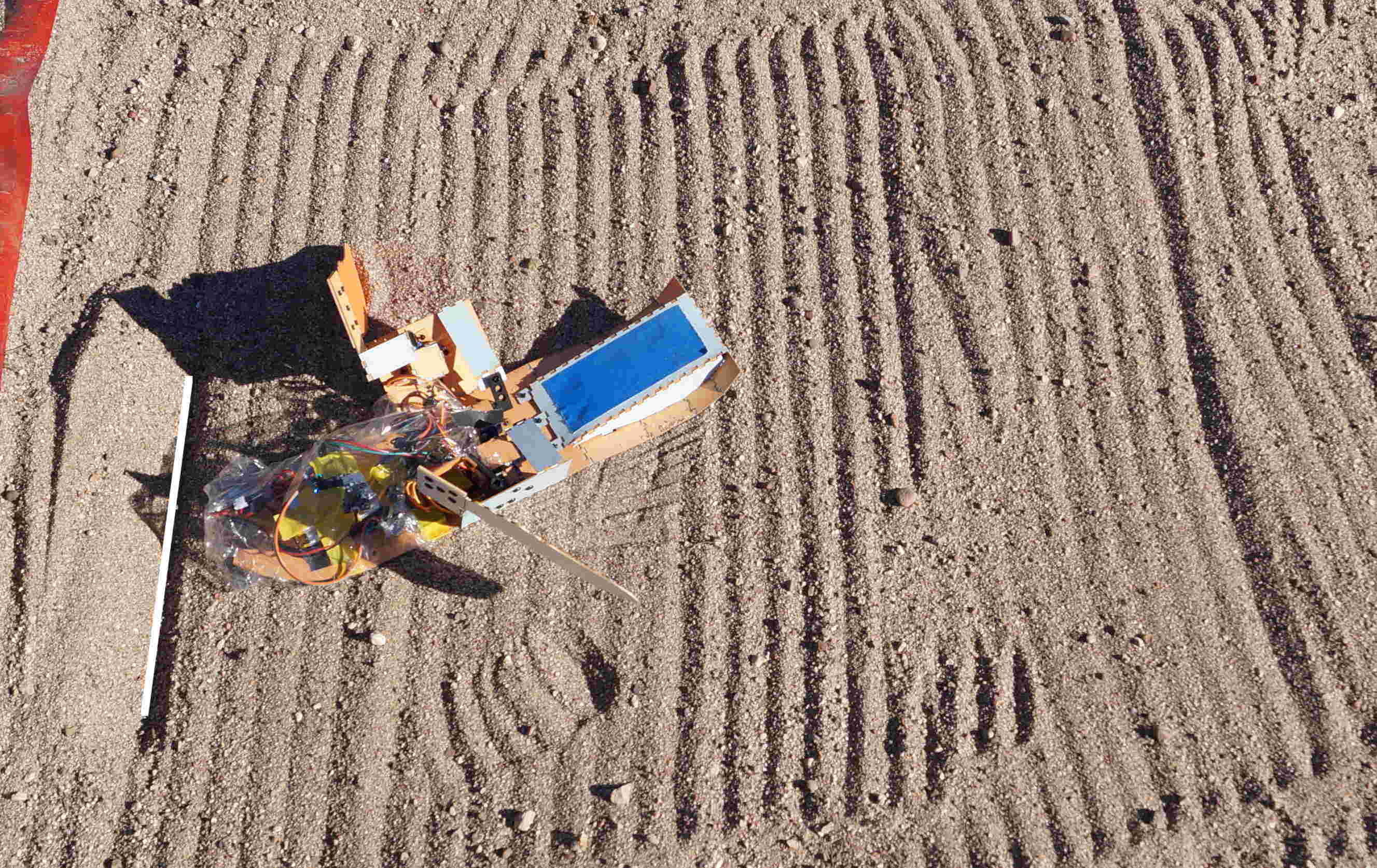}
        	\caption{Executions of policies learned on poppy seeds and executed in a real desert environment. The white line shows the start position of the robot.}
    	\end{subfigure}
        
        \begin{subfigure}[b]{\textwidth}
        	
        	\centering
        	\includegraphics[width=0.19\textwidth]{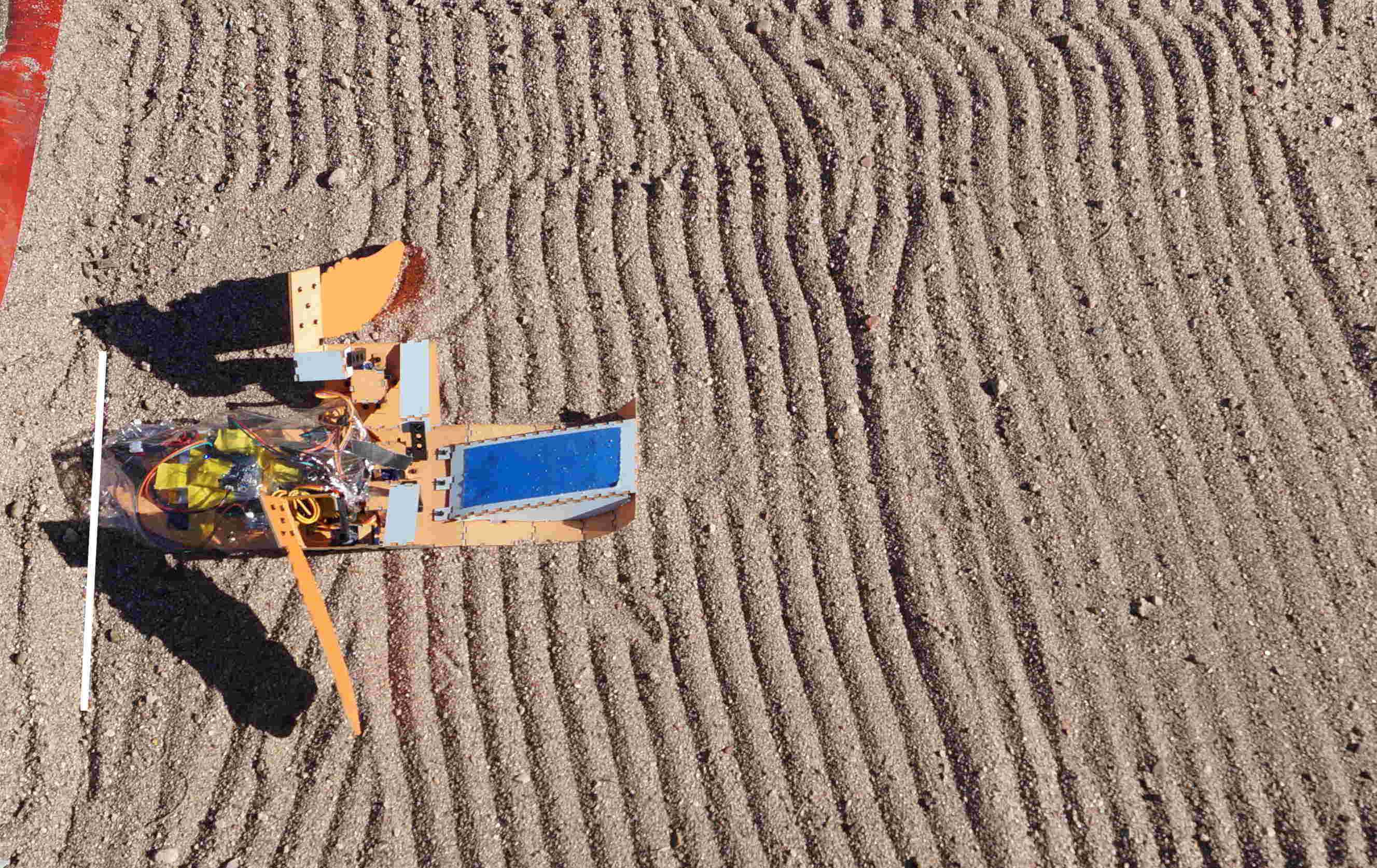}~%
            \includegraphics[width=0.19\textwidth]{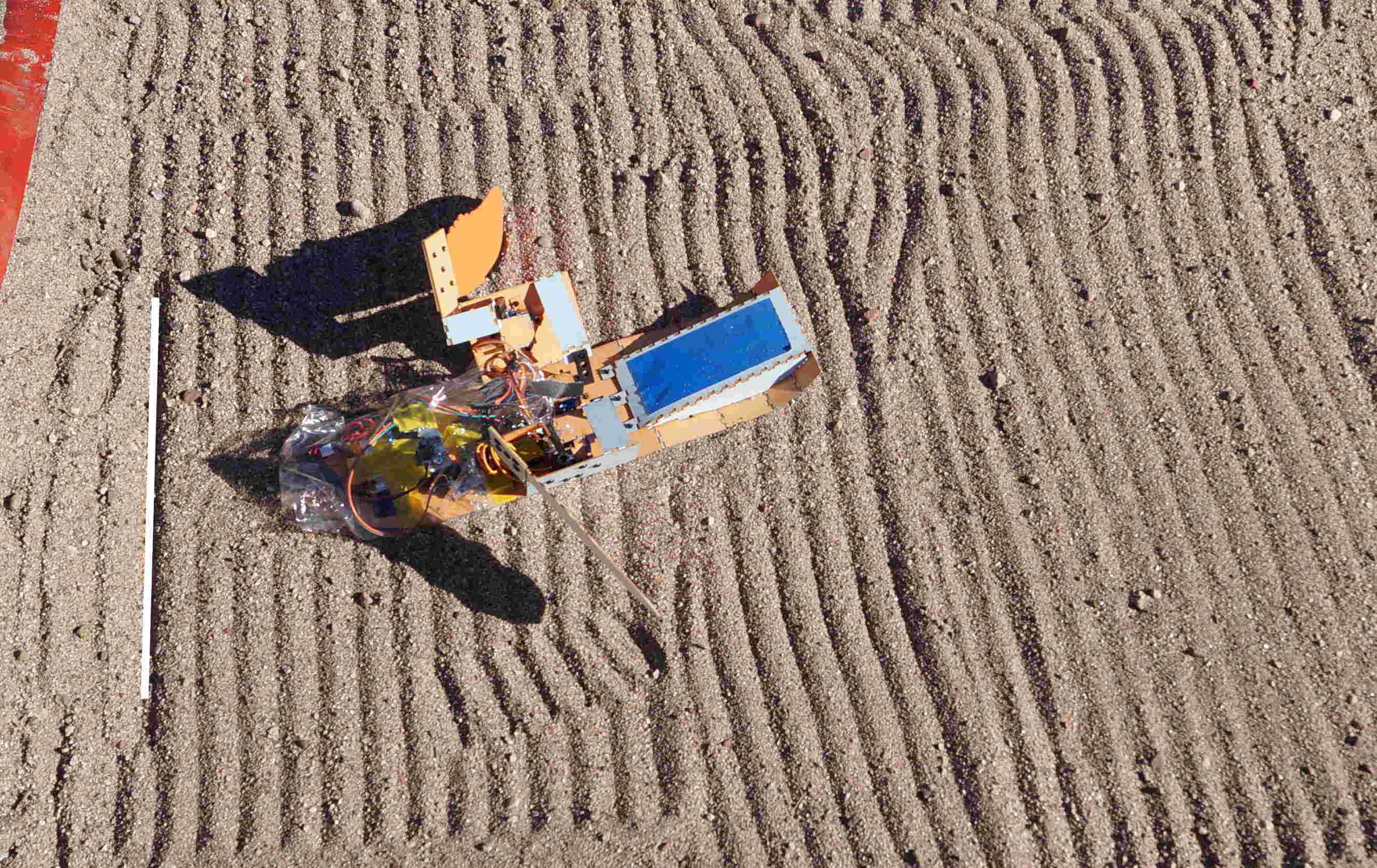}~%
            \includegraphics[width=0.19\textwidth]{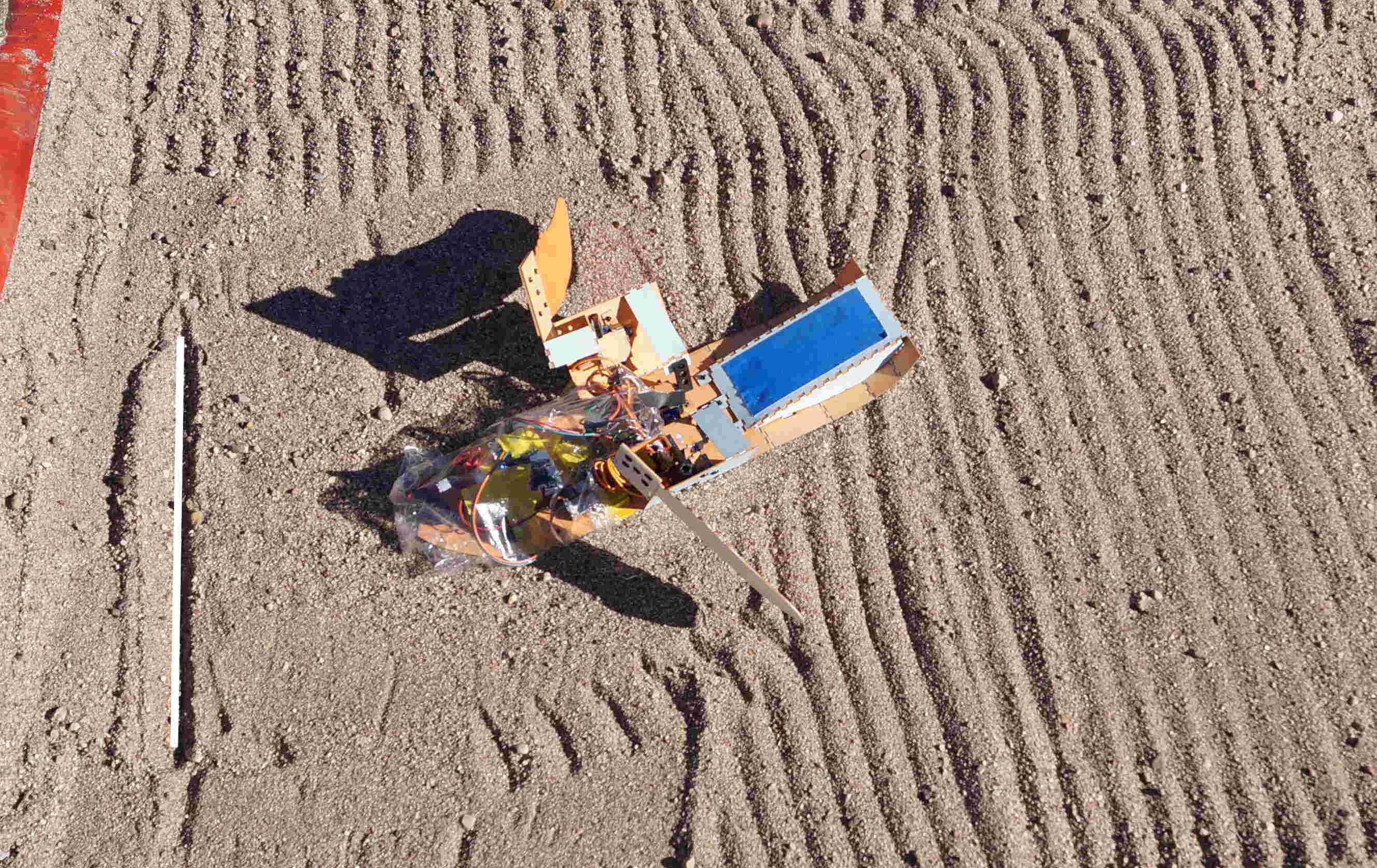}~%
            \includegraphics[width=0.19\textwidth]{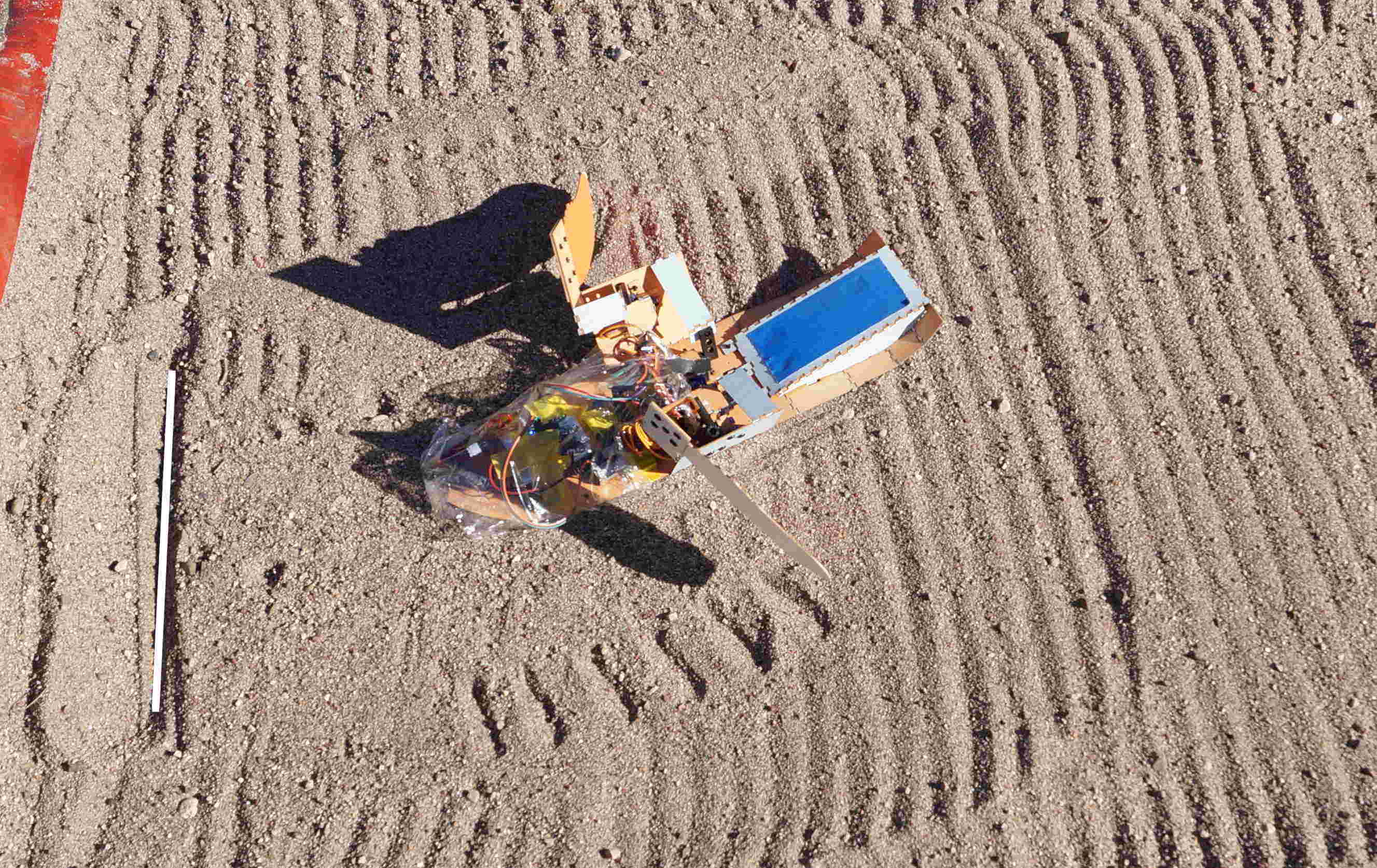}~%
            \includegraphics[width=0.19\textwidth]{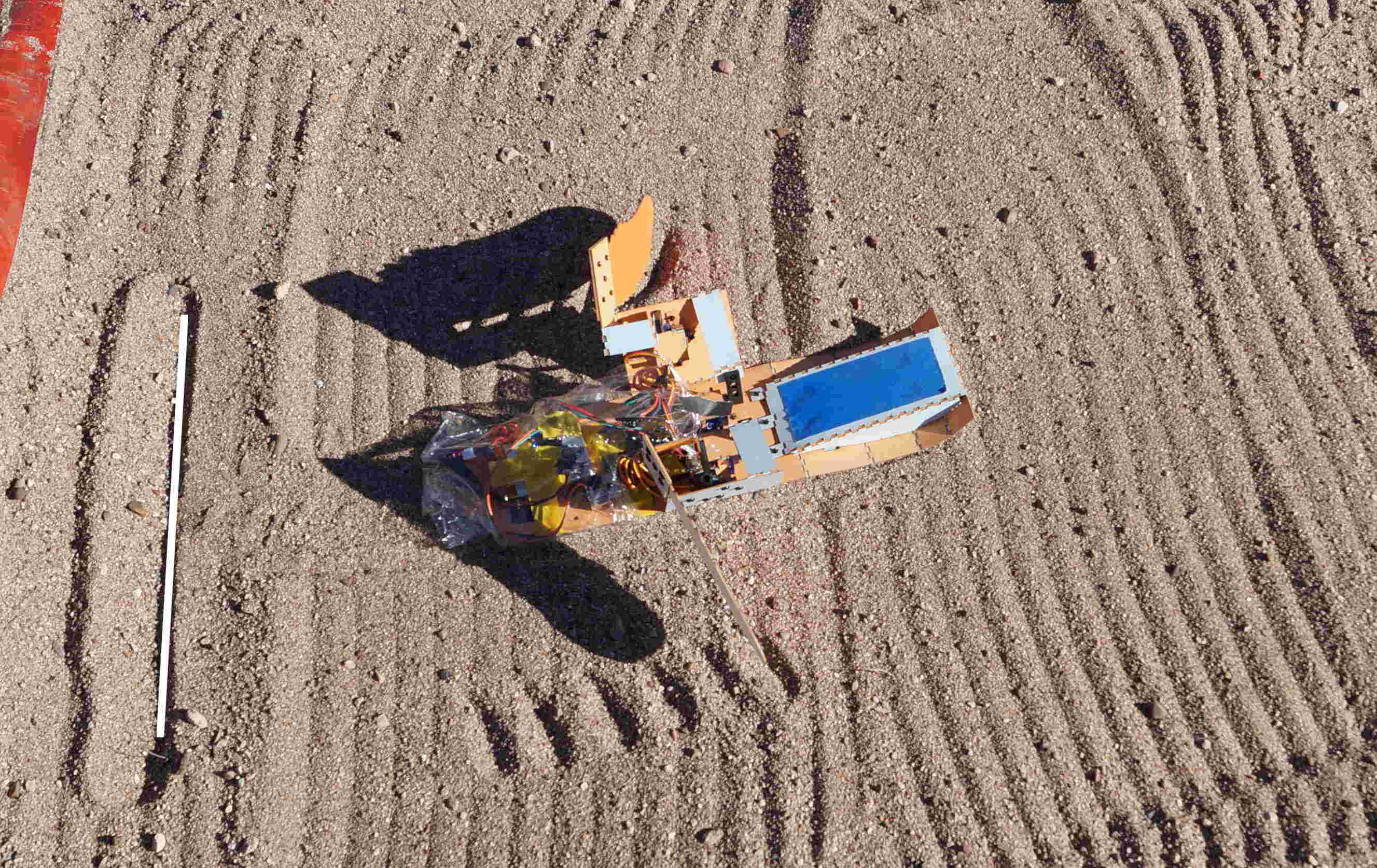}
        	\caption{Executions of policies learned in a desert environment. The white line shows the start position of the robot.\hspace*{26ex}}
    	\end{subfigure}
    \caption{Executions of learned policies on poppy seeds and in a real desert environment. Row (a) shows the execution of the policies learned on poppy seeds which are also executed in a real desert environment in (b). Finally, (c) shows the policies learned and executed in the desert. For both learning experiments the same initial values and random number generators were used. The images show the executions of trajectories after 1, 4, 6, 8 and 10 iterations.}
    \label{img::pictures::desert}
	\end{figure*}
    
    A series of images from the executions of the policies are shown in Figure \ref{img::pictures::desert}. The pictures show the final position after execution of policies learned in iteration one, four, six, eight and ten. The images in Figure \ref{img::pictures::desert} (b) and (c) show the difference in covered length between policies learned on poppy seeds and the policies learned in the natural environment.
    

\begin{figure}[htb]
	\centering
	\includegraphics[width=0.35\textwidth]{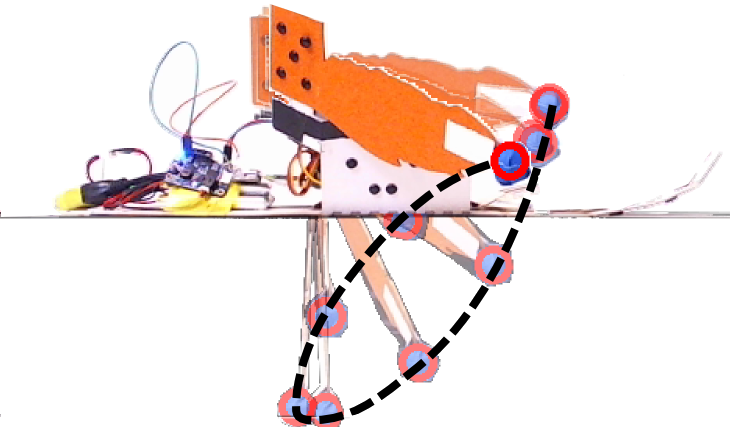}
	\includegraphics[width=0.36\textwidth]{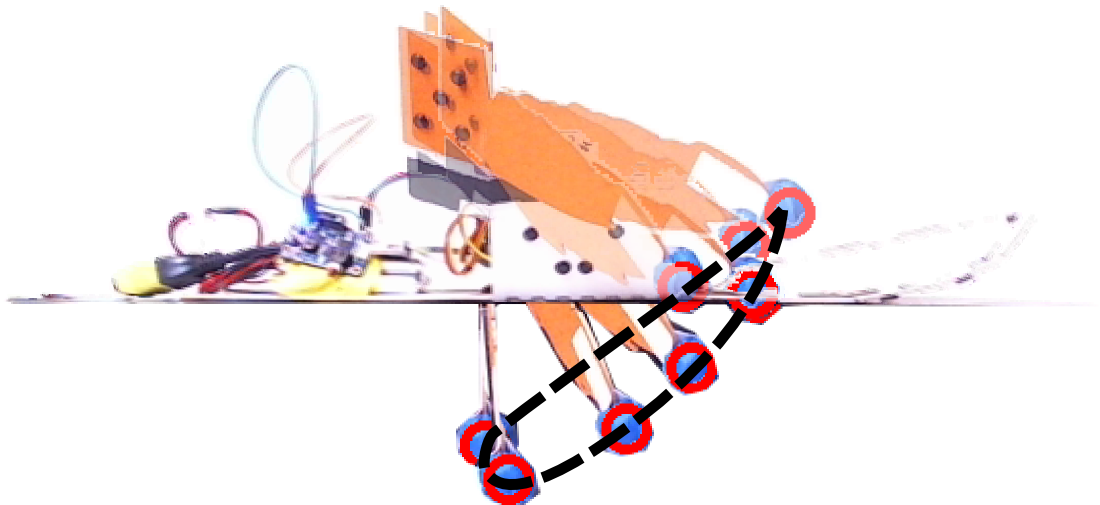}
	\caption{Top: the gait produced by the right fin after iteration 10 with fin A. Bottom: the gait produced by the right fin of the robot after iteration 3.}
	\label{img::sidestroke}
\end{figure}

%
%
\section{Discussion}



The results shown in Fig.~\ref{img::exp::exhaustive} and Fig.~\ref{img::exp::outdoor} indicate that for every fin that underwent learning, in both artificial and natural environments, the final locomotion policy shows some degree of improvement with regard to distance traveled by the robot after only 10 iterations.
This supports hypothesis H1 which postulated that Group Factor Policy Search would find an improved locomotion policy in a limited number of trials, despite variations in the environment and fin shape.

However, the results also indicate that some fins clearly performed better than others.
For example, fin B only achieved a mean pixel reward of 35.2 in the artificial environment, while fin A saw a mean pixel reward of 141.8, as shown in Fig.~\ref{img::exp::exhaustive::ab}.
This supports H2, which hypothesized that the physical shape of the fin affects locomotion performance.

It is interesting to note that the biologically inspired fins (A and C) out-performed the artificial fins (B and D) on average.
At least part of this may be due to the intersection of the fin and the ground when they make contact at an angle, as is the case in our robotic design.
The biological fins have a curved design which increases the surface area that is in contact with granulate media when compared to the artificial fins while the overall surface areas of artificial fins and biologically inspired fins are comparable to each other.
Furthermore, fin B exhibited significant deformation when in contact with the ground which likely reduced its effectiveness in producing forward motion.

The results shown in Fig.~\ref{img::exp::outdoor} support hypothesis H3, in that policies learned in the natural environment outperform the policies that were learned in the artificial environment.
We reason that part of this discrepancy is due to the composition of the granulate material.
The poppy seeds used in the artificial environment have an average density of 0.54 g/ml with a -- qualitatively speaking -- homogeneous seed size, while the sand grains in the desert have an average density of 1.46 g/ml and a heterogeneous grain size.
These results indicate that artificial environments consisting of popular granulate substitutes, such as poppy seeds, may not yield performance comparable to the real-world environments that they are mimicking.
Thus, it is not only simulations that can yield performance discrepancies, but also physical environments.

Additionally, we observed that the composition of the natural environment itself fluctuated over time.
For instance, we measured a difference in the moisture content of the sand of nearly 82\% between the two days in which we performed experiments: 1.59\% and 0.87\% by weight.
These factors may serve to make the target environment difficult to emulate, and suggest that not only are discrepancies possible between simulated environments, artificial environments, and actual environments, but also between the same actual environment over time.
We suspect that lifelong learning might be a possible solution to this problem. 

Yet another interesting observation can be made from the gaits shown in Fig.~\ref{img::sidestroke}.
The cycle produced by the fin during a more effective policy extends deeper and further than that produced during a less effective policy.
Intuitively, we can reason that this more effective policy pushes against a larger volume of sand, generating more force for forward motion.

\section{Conclusion}
In this paper, we presented a methodology for rapid prototyping of robotic structures for terrestrial locomotion. A combination of laminate robot manufacturing and sample-efficient reinforcement learning enables re-configuration and adaptation of both form and function to best fit environmental constraints. In turn, this approach decreases the amount of time for the development-production-learning-deployment cycle. With the presented techniques, it is possible to construct a robot out of raw material and learn a controller for locomotion in under a day. We designed a bio-inspired robotic device using the new methodology and, consequently, conducted an extensive robot learning study which involved several thousand executions. The experiment was performed with different sets of fins, both inside the lab, as well as in the desert of Arizona. Our results indicate the approach is well-suited for fast adaptation to new ground. 

The results also show that granulates which are commonly used as a replacement for sand in robotics laboratories may not be an effective replacement. More specifically, the efficiency of robot control policies learned on such granulates in the laboratory were not as effective when deployed outside. A variety of factors such as variability in actuation, energy supply, the manufacturing process, or the terrain may contribute to this phenomenon. Consequently, learning and adaptation is of crucial importance. The discussed sample-efficient reinforcement learning algorithm enabled robots to quickly adapt an existing policy or learn a new one. Learning time was typically in the range of $2-3$ hours. The results also show that biological inspiration in the fin design can lead to significant advantages in the resulting policies, even when learning was employed.  

For future work we aim to investigate life-long learning approaches that do not separate between a training and a deployment phase. Using an accelerometer, the robot could continuously calculate rewards and update the control policy. 

\FloatBarrier
\bibliographystyle{plainnat}
\bibliography{references,dan}

\begin{thebibliography}{26}
\providecommand{\natexlab}[1]{#1}
\providecommand{\url}[1]{\texttt{#1}}
\expandafter\ifx\csname urlstyle\endcsname\relax
  \providecommand{\doi}[1]{doi: #1}\else
  \providecommand{\doi}{doi: \begingroup \urlstyle{rm}\Url}\fi

\bibitem[Askari and Kamrin(2016)]{askari2016}
Hesam Askari and Ken Kamrin.
\newblock Intrusion rheology in grains and other flowable materials.
\newblock \emph{Nature Materials}, 15\penalty0 (12):\penalty0 1274--1279, 2016.

\bibitem[Bhushan(2009)]{bhushan2009biomimetics}
Bharat Bhushan.
\newblock Biomimetics: lessons from nature--an overview.
\newblock \emph{Philosophical Transactions of the Royal Society A:
  Mathematical, Physical and Engineering Sciences}, 367\penalty0
  (1893):\penalty0 1445--1486, 2009.

\bibitem[Clark et~al.(2001)Clark, Cham, Bailey, Froehlich, Nahata, Full, and
  Cutkosky]{clark2001biomimetic}
Jonathan~E Clark, Jorge~G Cham, Sean~A Bailey, Edward~M Froehlich, Pratik~K
  Nahata, Robert~J Full, and Mark~R Cutkosky.
\newblock Biomimetic design and fabrication of a hexapedal running robot.
\newblock In \emph{Proceedings of IEEE International Conference on Robotics and
  Automation}, volume~4, pages 3643--3649, 2001.

\bibitem[Dodd~Jr(1988)]{dodd1988}
C~Kenneth Dodd~Jr.
\newblock Synopsis of the biological data on the loggerhead sea turtle caretta
  caretta (linnaeus 1758).
\newblock Technical report, DTIC Document, 1988.

\bibitem[Eckert and Luginbuhl(1988)]{eckert1988}
Karen~L Eckert and Chris Luginbuhl.
\newblock Death of a giant.
\newblock \emph{Marine Turtle Newsletter}, 43:\penalty0 2--3, 1988.

\bibitem[Gollnick et~al.(2011)Gollnick, Magleby, and Howell]{Gollnick2011}
Paul~S. Gollnick, Spencer~P. Magleby, and Larry~L. Howell.
\newblock {An Introduction to Multilayer Lamina Emergent Mechanisms}.
\newblock \emph{Journal of Mechanical Design}, 133\penalty0 (8):\penalty0
  081006, 2011.
\newblock ISSN 10500472.

\bibitem[Holmes et~al.(2006)Holmes, Full, Koditschek, and
  Guckenheimer]{holAful06}
Philip Holmes, Robert~J Full, Dan Koditschek, and John Guckenheimer.
\newblock The dynamics of legged locomotion: Models, analyses, and challenges.
\newblock \emph{Siam Review}, 48\penalty0 (2):\penalty0 207--304, 2006.

\bibitem[Krouchev et~al.(2006)Krouchev, Kalaska, and Drew]{Krouchev1991}
Nedialko Krouchev, John~F. Kalaska, and Trevor Drew.
\newblock Sequential activation of muscle synergies during locomotion in the
  intact cat as revealed by cluster analysis and direct decomposition.
\newblock \emph{Journal of Neurophysiology}, 96\penalty0 (4):\penalty0
  1991--2010, 2006.
\newblock ISSN 0022-3077.

\bibitem[Li et~al.(2013)Li, Zhang, and Goldman]{li2013}
Chen Li, Tingnan Zhang, and Daniel~I. Goldman.
\newblock A terradynamics of legged locomotion on granular media.
\newblock \emph{Science}, 339:\penalty0 1408--1412, 2013.

\bibitem[Low et~al.(2007)Low, Zhou, Ong, and Yu]{low2007modular}
Kin-Huat Low, Chunlin Zhou, TW~Ong, and Junzhi Yu.
\newblock Modular design and initial gait study of an amphibian robotic turtle.
\newblock In \emph{Robotics and Biomimetics, 2007. ROBIO 2007. IEEE
  International Conference on}, pages 535--540. IEEE, 2007.

\bibitem[Luck et~al.(2016)Luck, Pajarinen, Berger, Kyrki, and
  Amor]{luck2016sparse}
Kevin~Sebastian Luck, Joni Pajarinen, Erik Berger, Ville Kyrki, and Heni~Ben
  Amor.
\newblock Sparse latent space policy search.
\newblock In \emph{AAAI}, pages 1911--1918, 2016.

\bibitem[Ma et~al.(2013)Ma, Chirarattananon, Fuller, and
  Wood]{ma2013controlled}
Kevin~Y Ma, Pakpong Chirarattananon, Sawyer~B Fuller, and Robert~J Wood.
\newblock Controlled flight of a biologically inspired, insect-scale robot.
\newblock \emph{Science}, 340\penalty0 (6132):\penalty0 603--607, 2013.

\bibitem[Maladen et~al.(2009)Maladen, Ding, Li, and
  Goldman]{maladen2009undulatory}
Ryan~D Maladen, Yang Ding, Chen Li, and Daniel~I Goldman.
\newblock Undulatory swimming in sand: subsurface locomotion of the sandfish
  lizard.
\newblock \emph{science}, 325\penalty0 (5938):\penalty0 314--318, 2009.

\bibitem[Mazouchova et~al.(2010)Mazouchova, Gravish, Savu, and
  Goldman]{mazouchova2010}
Nicole Mazouchova, Nick Gravish, Andrei Savu, and Daniel~I Goldman.
\newblock Utilization of granular solidification during terrestrial locomotion
  of hatchling sea turtles.
\newblock \emph{Biology Letters}, 6:\penalty0 398--401, 2010.

\bibitem[Mazouchova et~al.(2013{\natexlab{a}})Mazouchova, Umbanhowar, and
  Goldman]{mazouchova2013}
Nicole Mazouchova, Paul~B Umbanhowar, and Daniel~I Goldman.
\newblock Flipper-driven terrestrial locomotion of a sea turtle-inspired robot.
\newblock \emph{Bioinspiration and Biomimetics}, 8\penalty0 (2):\penalty0
  026007, 2013{\natexlab{a}}.

\bibitem[Mazouchova et~al.(2013{\natexlab{b}})Mazouchova, Umbanhowar, and
  Goldman]{mazouchova2013flipper}
Nicole Mazouchova, Paul~B Umbanhowar, and Daniel~I Goldman.
\newblock Flipper-driven terrestrial locomotion of a sea turtle-inspired robot.
\newblock \emph{Bioinspiration \& biomimetics}, 8\penalty0 (2):\penalty0
  026007, 2013{\natexlab{b}}.

\bibitem[Mnih et~al.(2015)Mnih, Kavukcuoglu, Silver, Rusu, Veness, Bellemare,
  Graves, Riedmiller, Fidjeland, Ostrovski, Petersen, Beattie, Sadik,
  Antonoglou, King, Kumaran, Wierstra, Legg, and Hassabis]{mnih-dqn-2015}
Volodymyr Mnih, Koray Kavukcuoglu, David Silver, Andrei~A. Rusu, Joel Veness,
  Marc~G. Bellemare, Alex Graves, Martin Riedmiller, Andreas~K. Fidjeland,
  Georg Ostrovski, Stig Petersen, Charles Beattie, Amir Sadik, Ioannis
  Antonoglou, Helen King, Dharshan Kumaran, Daan Wierstra, Shane Legg, and
  Demis Hassabis.
\newblock Human-level control through deep reinforcement learning.
\newblock \emph{Nature}, 518\penalty0 (7540):\penalty0 529--533, 02 2015.

\bibitem[Playter et~al.(2006)Playter, Buehler, and Raibert]{plaAbue06}
Robert Playter, Martin Buehler, and Marc Raibert.
\newblock Bigdog.
\newblock In Douglas W.~Gage Grant R.~Gerhart, Charles M.~Shoemaker, editor,
  \emph{Unmanned Ground Vehicle Technology VIII}, volume 6230 of
  \emph{Proceedings of SPIE}, pages 62302O1--62302O6, 2006.

\bibitem[Pritchard and Mortimer(1999)]{pritchard1999}
Peter Pritchard and Jeanne Mortimer.
\newblock Taxonomy, external morphology, and species identification.
\newblock \emph{Research and management techniques for the conservation of sea
  turtles}, 4:\penalty0 21, 1999.

\bibitem[Sreetharan et~al.(2012)Sreetharan, Whitney, Strauss, and
  Wood]{Sreetharan2012}
Pratheev~S Sreetharan, John~P Whitney, Mark~D Strauss, and Robert~J Wood.
\newblock {Monolithic fabrication of millimeter-scale machines}.
\newblock \emph{Journal of Micromechanics and Microengineering}, 22\penalty0
  (5):\penalty0 055027, may 2012.
\newblock ISSN 0960-1317.

\bibitem[Sutton and Barto(1998)]{Sutton:1998:IRL:551283}
Richard~S. Sutton and Andrew~G. Barto.
\newblock \emph{Introduction to Reinforcement Learning}.
\newblock MIT Press, Cambridge, MA, USA, 1st edition, 1998.
\newblock ISBN 0262193981.

\bibitem[Tesch et~al.(2009)Tesch, Lipkin, Brown, Hatton, Peck, Rembisz, and
  Choset]{tesch2009parameterized}
Matthew Tesch, Kevin Lipkin, Isaac Brown, Ross Hatton, Aaron Peck, Justine
  Rembisz, and Howie Choset.
\newblock Parameterized and scripted gaits for modular snake robots.
\newblock \emph{Advanced Robotics}, 23\penalty0 (9):\penalty0 1131--1158, 2009.

\bibitem[Whitney et~al.(2011)Whitney, Sreetharan, Ma, and Wood]{Whitney2011}
John~P Whitney, Pratheev~S Sreetharan, Kevin~Y Ma, and Robert~J Wood.
\newblock {Pop-up book MEMS}.
\newblock \emph{Journal of Micromechanics and Microengineering}, 21\penalty0
  (11):\penalty0 115021, nov 2011.
\newblock ISSN 0960-1317.

\bibitem[Wood et~al.(2008)Wood, Avadhanula, Sahai, Steltz, and
  Fearing]{Wood2008}
Robert~J Wood, Srinath Avadhanula, Ranjana Sahai, Erik Steltz, and Ronald~S
  Fearing.
\newblock {Microrobot Design Using Fiber Reinforced Composites}.
\newblock \emph{Journal of Mechanical Design}, 130\penalty0 (5):\penalty0
  052304, 2008.
\newblock ISSN 10500472.

\bibitem[Wyneken(1997)]{wyneken1997}
Jeanette Wyneken.
\newblock Sea turtle locomotion: Mechanics, behavior, and energetics.
\newblock In Peter~L Lutz, editor, \emph{The Biology of Sea Turtles}, pages
  168--198. CRC Press, 1997.

\bibitem[Yao et~al.(2013)Yao, Liang, Wang, Yang, Shen, Zhang, Wu, and
  Tian]{yao2013development}
Guocai Yao, Jianhong Liang, Tianmiao Wang, Xingbang Yang, Qi~Shen, Yucheng
  Zhang, Hailiang Wu, and Weicheng Tian.
\newblock Development of a turtle-like underwater vehicle using central pattern
  generator.
\newblock In \emph{Robotics and Biomimetics (ROBIO), 2013 IEEE International
  Conference on}, pages 44--49. IEEE, 2013.

\end{thebibliography}

\end{document}